\pdfoutput=1

\documentclass[11pt]{article}

\usepackage[final]{acl}
\usepackage{tcolorbox}
\tcbuselibrary{skins, breakable}
\newtcolorbox[auto counter, number within=section]{promptbox}[2][]{%
    colframe=blue!75!black,
    colback=blue!10,
    coltitle=white,
    fonttitle=\small\bfseries,
    title=Prompt Template~\thetcbcounter: #2,
    breakable, 
    enhanced,
    fontupper=\ttfamily,
    #1 
}

\usepackage{amsmath}
\usepackage{times}
\usepackage{latexsym}
\usepackage{booktabs}
\usepackage{multirow}
\usepackage[table,xcdraw]{xcolor}
\usepackage[T1]{fontenc}

\usepackage[utf8]{inputenc}

\usepackage{microtype}

\usepackage{inconsolata}

\usepackage{graphicx}

\title{\textsc{LogiDynamics}: Unraveling the Dynamics of Inductive, Abductive and Deductive Logical Inferences in LLM Reasoning}

\author{Tianshi Zheng$^{\spadesuit}$,  Jiayang Cheng$^{\spadesuit}$, Chunyang Li$^{\spadesuit}$, Haochen Shi$^{\spadesuit}$, Zihao Wang$^{\spadesuit}$, Jiaxin Bai$^{\spadesuit}$\\ \textbf{Yangqiu Song}$^{\spadesuit}$, \textbf{Ginny Y. Wong}$^{\clubsuit}$, \textbf{Simon See}$^{\clubsuit}$ \\
  $^{\spadesuit}$Department of Computer Science and Engineering, HKUST, Hong Kong SAR, China\\
  $^{\clubsuit}$NVIDIA AI Technology Center (NVAITC), NVIDIA, Santa Clara, USA\\
  \texttt{tzhengad@connect.ust.hk}\\
}

\begin{document}
\maketitle
\begin{abstract}
Modern large language models (LLMs) employ diverse logical inference mechanisms for reasoning, making the strategic optimization of these approaches critical for advancing their capabilities. This paper systematically investigate the \textbf{comparative dynamics} of inductive (System 1) versus abductive/deductive (System 2) inference in LLMs. We utilize a controlled analogical reasoning environment\footnote{ \href{https://github.com/HKUST-KnowComp/LogiDynamics}{https://github.com/HKUST-KnowComp/LogiDynamics}}, varying modality (textual, visual, symbolic), difficulty, and task format (MCQ / free-text). Our analysis reveals System 2 pipelines generally excel, particularly in visual/symbolic modalities and harder tasks, while System 1 is competitive for textual and easier problems. Crucially, task format significantly influences their relative advantage, with System 1 sometimes outperforming System 2 in free-text rule-execution. These core findings generalize to broader in-context learning. Furthermore, we demonstrate that advanced System 2 strategies like hypothesis selection and iterative refinement can substantially scale LLM reasoning. This study offers foundational insights and actionable guidelines for strategically deploying logical inference to enhance LLM reasoning. 
\end{abstract}

\begin{quote}
\centering
\emph{"It is not enough to have a good mind;\\ the main thing is to use it well."} \\[1ex]
{\raggedleft\small\textemdash\ René Descartes\par}

\end{quote}

\begin{figure}[t]
\begin{center}
\includegraphics[clip,width=220pt]{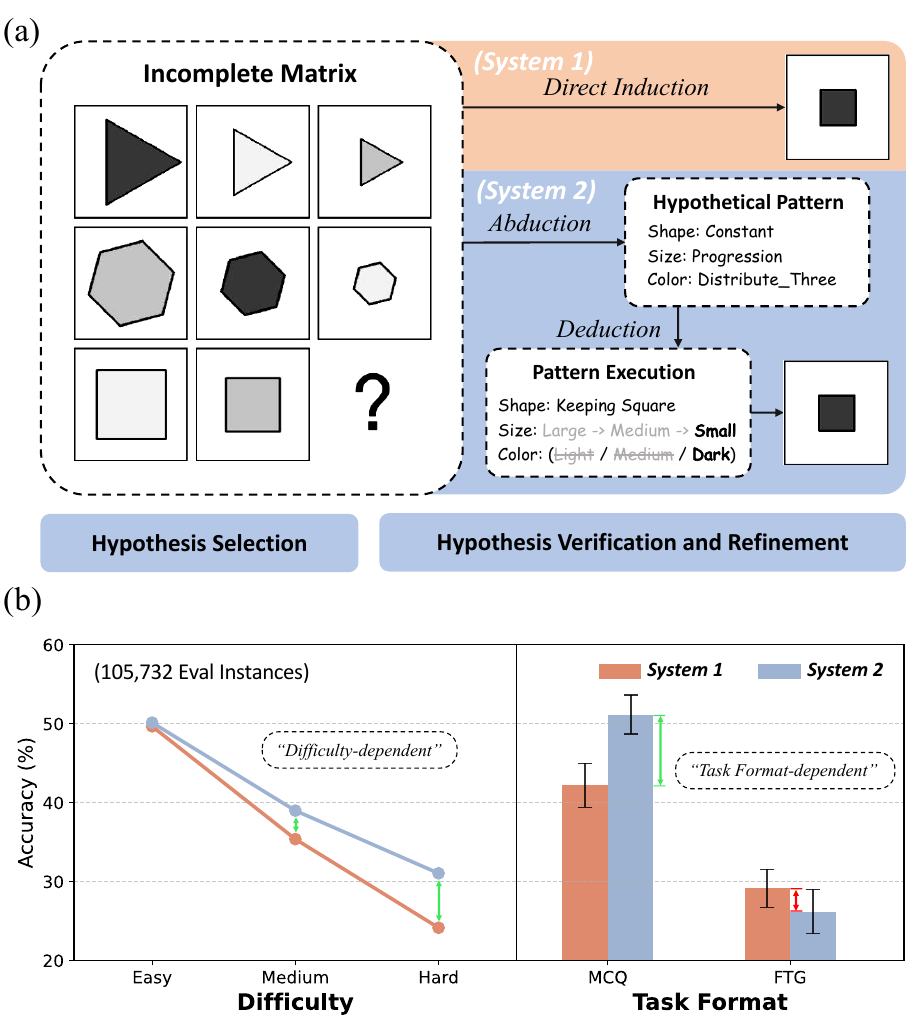}
\end{center}

\caption{(a) An illustration of System 1 and System 2 logical inference pipelines in RAVEN's progressive matrix. (b) General \textbf{comparative dynamics} between System 1 and System 2 pipelines in all experiments.}

\label{fig:raven}
\end{figure}

\section{Introduction}
Logical Inference\footnote{The term `inference' encompasses multiple interpretations across different disciplines. This paper employs the term strictly within its logical trichotomy: deductive, inductive, and abductive inference, as defined in \cite{flach2000abduction}.} is the reasoning process of deriving conclusions from known premises \cite{copi1990introduction,deductivereasoningjohnson}. It primarily categorizes into \textit{deductive inference} --- where conclusions follow with logical necessity from premises, and \textit{inductive inference} --- where conclusions serves as general rules derived from specific instances \cite{salmon_1984}. While the introduction of \textit{abductive inference} \cite{peirce_1958, frankfurt1958peirce} serves as a third perspective, denoting the process of forming an explanatory hypothesis from an observation requiring explanation. Logical inference plays a crucial role in artificial intelligence, scientific research, and philosophy, where rational decision-making and hypothesis formation are foundational \cite{Hempel_Oppenheim_1948,684604fc-4b8d-3190-bf8c-948758999eb7,REITER198757}.

Different logical inference pipelines can be applied in solving the same reasoning task. Figure \ref{fig:raven}(a) illustrates an example of Raven's Progressive Matrices \cite{raven1938,zhang2019raven}, where the missing element in the 3×3 matrix is inferred through the common patterns among different rows. There are two approaches to solving this problem: 1) directly inferring the missing element from the observed elements in the matrix, and 2) explicitly identifying the common patterns across rows, then deductively applying these patterns to determine the missing element in the last row. The former is driven by inductive inference and features fast, intuitive, pattern-recognition guided reasoning. The latter consists of abductive and deductive inference, featuring slower but more deliberate analysis. These approaches correspond to System 1 and System 2 thinking, respectively \cite{kahneman2011thinking}.

Research on large language models (LLMs) has explored the logical inference pipelines employed by LLMs for solving a wide range of tasks. \citet{qiu2024phenomenal} and \citet{wang2024hypothesis} have demonstrated the effectiveness of the System 2 approach in various inductive reasoning datasets such as ARC \cite{chollet2019measure} and its variants \cite{kim2022playgrounds,xu2023llms}. \citet{he2024ideaenhancingrulelearning} highlighted the potential of System 2 logical inference in the reasoning workflow of LLM-based agents. While \citet{liu2024incompleteloopinstructioninference} compared both System 1 and System 2 approaches in several in-context learning tasks, pointing out the inconsistency of their relative performances across datasets. Nevertheless, all prior studies leave an open question: \textit{When} and \textit{how} can System 1 and System 2 logical inference pipelines be effectively leveraged to enhance LLM reasoning?

To address this intricate question, we systematically investigate the \textbf{comparative dynamics} of System 1 and System 2 pipelines within LLM reasoning tasks, specifically examining the contingency of their performance preferences on task attributes such as modality, difficulty, and task format. First, we build a fully controllable evaluation environment using analogical reasoning tasks. The environment is controlled in three dimensions: 1) \textit{Modality}: The data covers textual (word/phrase), visual (images), and symbolic modalities. 2) \textit{Difficulty}: All tasks are labeled with relative difficulty levels (easy, medium, and hard). 3) \textit{Task Format}: For each question, we provide two task formats: multiple-choice questions (MCQ) or free-text generation (FTG) format.

With experiments in 10 modern LLMs (and MLLMs), \textbf{we discover several key findings}:
\begin{itemize}
    \item \textbf{Modality-dependent}: System 2 logical inference shows superior performance in \textit{visual} and \textit{symbolic} tasks, while System 1 performs comparably in \textit{textual} tasks.
    \item \textbf{Difficulty-dependent}: System 2 logical inference is more advantageous in \textit{harder} tasks, while System 1 achieve comparable performance in \textit{easier} tasks.
    \item \textbf{Task Format-dependent}: For tasks involving explicit rule execution, System 1 logical inference outperforms System 2 in \textit{FTG} format, but underperforms in \textit{MCQ} format.
\end{itemize}

To verify the generalizability of our findings, we conduct further experiments in the List Function dataset \cite{rule2020child} and SALT dataset (ours), where we observe similar comparative dynamics in difficulty and task format. We argue that \textbf{our findings can be generalized to broader in-context learning (ICL) tasks} where: 1) the few-shot demonstrations are presented in Input-Output format, and 2) the mapping function between input and output can be explicitly defined.

Furthermore, we explored the effects of more sophisticated System 2 logical inference pipelines, including hypothesis selection, hypothesis verification, and refinement. Using these paradigms, LLMs demonstrate significant performance improvements as the number of inference tokens increases. We show that, with sufficient computational resources, \textbf{LLMs under logical inference scaling achieve performance comparable to state-of-the-art Long-CoT reasoning models}. This highlights the potential of scaling inference through advanced System 2 logical inference pipelines.

This work makes several key contributions to understanding and improving LLM reasoning capabilities from a logical inference perspective:
\begin{enumerate}
\item We provide a \textbf{systematic evaluation environment} to compare logical inference paradigms across controlled dimensions.\,(\S\ref{sec:eval}) 
\item We present \textbf{rich findings as clear guidelines} for leveraging different inference approaches based on task characteristics.\,(\S\ref{sec:main}) 
\item We validate our findings' \textbf{generalizability to broader in-context learning tasks}.\,(\S\ref{sec:generalization}) 
\item We highlight the potential to \textbf{scale up LLM reasoning} using advanced System 2 logical inference paradigms.\,(\S\ref{sec:scaling}) 
\end{enumerate}
Collectively, these contributions establish a foundation for future research on enhancing LLM reasoning through optimized logical inference strategies.

\section{Preliminaries}

\subsection{Analogical Reasoning}


Analogical reasoning is a fundamental aspect of cognitive intelligence \cite{gentner2001analogical}. It involves inferring a missing element in a target domain according to relational structures from a source domain. 
Formally, given a source pair \((A, A')\) and an incomplete target pair \((B, x)\), where \(A\) and \(A'\) have an implicit relational pattern \(P\), the goal is to infer \(x\) that have the same relational pattern \(P\) with \(B\). 
This task can be defined as:
\[
B' = \arg\max_{x \in \mathcal{X}} \text{sim}_P((A, A'), (B, x)),
\]
where \(\text{sim}_P\) measures the consistency of the relational pattern \(P\) between the source pair \((A, A')\) and the candidate target pair \((B, x)\), and \(\mathcal{X}\) represents the set of all possible candidates for \(B'\). The complete analogy is denoted as \(A : A' :: B : B'\).  For instance, given the source pair \((\textit{sun}, \textit{planet})\) and the incomplete target pair \((\textit{nucleus}, x)\), we can infer \(x = \textit{electron}\) by identifying the pattern \(P\) as \textit{orbital relationship}.


The task of analogical reasoning is particularly well-suited for our investigation for several reasons: 1) it offers a well-defined task structure while encompassing diverse data modalities, 2) it is compatible with a variety of logical inference pipelines, and 3) it is considered out-of-distribution for the training data of LLMs, enabling a robust evaluation of their reasoning capabilities under generalization \cite{claire2024analogy}.

\subsection{Logical Inference Pipelines}
In the main experiment, we compare three logical inference pipelines: direct induction, abduction + deduction, and automatic inference. More sophisticated pipelines involving hypothesis selection, verification, and refinement are discussed in the scaling experiments in Section \ref{sec:scaling}. Detailed prompt templates are provided in Appendix \ref{app:prompt}.
\paragraph{Direct Answering as Inductive Inference}
Inductive inference is often associated with fast, intuitive reasoning in cognition \cite{2af2fe0f-5f65-3dca-bfa3-cd96adbc1502}. Similar to \citet{liu2024incompleteloopinstructioninference}, we regard the direct answering of LLMs as a form of inductive inference, representing their System 1 logical inference pipeline.
\paragraph{Abductive and Deductive Inference} With this System 2 pipeline, task completion is decomposed into two steps. First, LLMs are required to abductively infer the hypothetical pattern \(P_h\) based on the source pair(s). Then, they deductively apply this pattern to the incomplete target pair as \( B \xrightarrow{P_h} B' \).
\paragraph{Zero-shot CoT as Automatic Inference} The reasoning process observed in zero-shot CoT (Chain-of-Thought) \cite{wei2023chainofthoughtpromptingelicitsreasoning}, which we term ``Automatic Inference'' for the purpose of this paper, demonstrates an inherent logical inference capability acquired during instruction-tuning or alignment stages. Therefore, we included the ``Automatic Inference'' in our comparison for reference.

\section{Evaluation Environment}
In this section, we introduce our evaluation environment of analogical reasoning, providing details on the settings for each control dimensions.

\label{sec:eval}
\begin{table}[]
\setlength{\tabcolsep}{3.5pt}
\centering
\scriptsize
\begin{tabular}{cllcccc}
\toprule
\multicolumn{3}{c}{\textbf{Dataset}} & \multicolumn{3}{c}{\textbf{Difficulty}} & \multirow{2}{*}{\textbf{Total}} \\ \cmidrule(lr){1-3}\cmidrule(lr){4-6}
Task & \multicolumn{1}{c}{Modality} & \multicolumn{1}{c}{Benchmark} & Easy & Medium & Hard &  \\ \midrule
\multirow{3}{*}{Analogy} & Textual & E-KAR & 317 & 435 & 496 & 1248 \\
 & Visual & VASR & 455 & 572 & 320 & 1347 \\
 & Symbolic & RAVEN & 402 & 462 & 395 & 1259 \\ \midrule
 \multirow{2}{*}{General ICL} & Math/Code & List Function & 432 & 423 & 395 & 1250 \\
 & Textual & SALT & 400 & 400 & 400 & 1200 \\ \midrule
\multicolumn{3}{c}{\textbf{Total}} & 2006 & 2292 & 2006 & \textbf{6304} \\ \bottomrule
\end{tabular}
\caption{Dataset statistics across modalities and difficulty levels. Details of general in-context learning tasks (List Function and SALT) are introduced in Section \ref{sec:generalization}.}
\label{tab:dataset_stats}

\end{table}
\begin{figure*}[t]
\begin{center}
\includegraphics[clip,width=\linewidth]{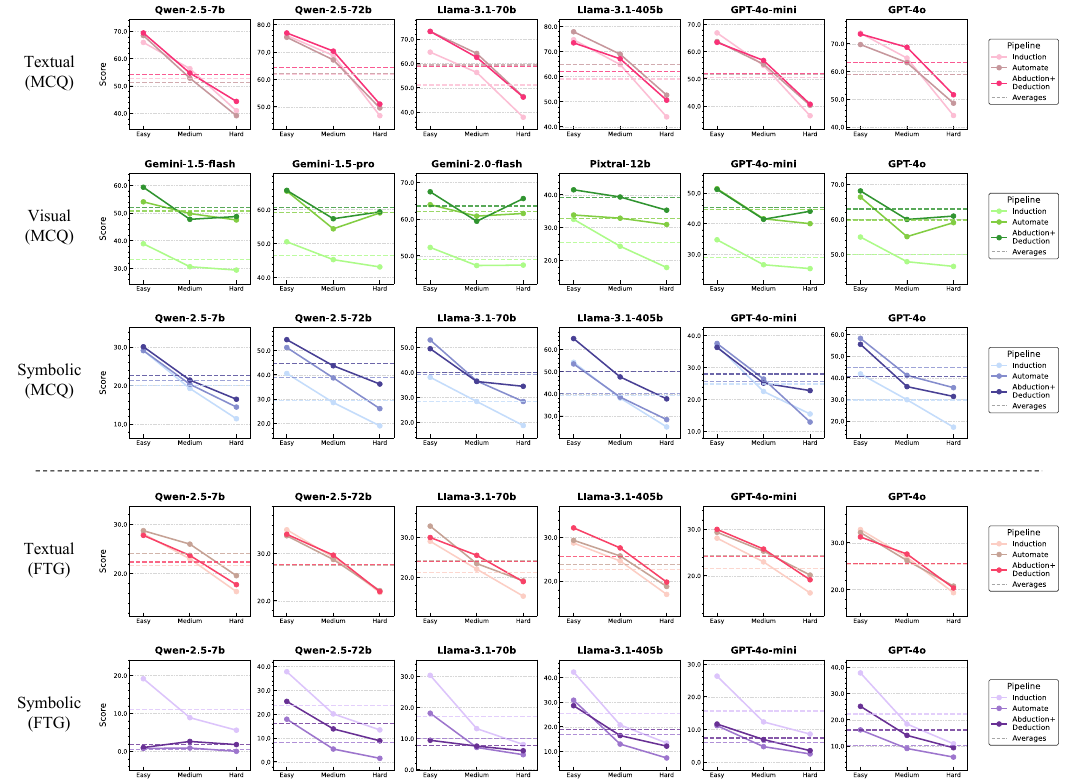}
\end{center}
\caption{LLM performances (in Accuracy \%) in our evaluation environment under different reasoning pipelines.}
\vspace{-0.3cm}
\label{fig:main_result}
\end{figure*}
\subsection{Modality}
Exploring diverse data modalities is crucial for obtaining comprehensive insights. To this end, we selected three analogical reasoning tasks across different modalities. \textbf{E-KAR} \cite{Chen_2022} consists of human-curated analogy questions between word pairs (or sets), where analogies are determined by shared ontological relationships between words. \textbf{VASR} \cite{bitton2022vasrvisualanalogiessituation} comprises human-annotated analogical questions between image pairs, where analogies are determined by shared semantic transitions between images. \textbf{RAVEN} \cite{raven1938,zhang2019raven,hu2022stratifiedruleawarenetworkabstract} generates symbolic matrices using attributed stochastic image grammar (A-SIG), where analogies are determined by shared attribute shifts among rows. To enhance comprehension in large language models, we adopt the abstracted version proposed by \citet{hu-etal-2023-context}, which tokenizes the matrix images into symbolic vectors.

\subsection{Difficulty}
Task difficulty, while a key determinant of thinking styles \cite{phillips2016thinking}, is largely overlooked in research on reasoning paradigms in LLMs. To address this, we conducted difficulty annotations for all three datasets. In analogical reasoning involving real-world data, difficulty is often measured by the semantic distance between analogy pairs \cite{Vendetti2012,Jones2022}. For E-KAR, we compute the semantic distance between word pairs using \textbf{FastText} embeddings \cite{bojanowski2017enriching}, which are more suitable than Word2Vec \cite{mikolov2013efficientestimationwordrepresentations} or BERT \cite{devlin2019bertpretrainingdeepbidirectional}, as the word pairs exhibit morphological variations but lack contextual dependencies. For VASR, we calculate the distance between \textbf{VGG} encodings \cite{simonyan2015deepconvolutionalnetworkslargescale} to account for both semantic and graphical features. For RAVEN, task complexity is defined by the number of \textbf{attribute variations} across the columns. The statistics of our datasets across different modalities and difficulty levels are presented in Table \ref{tab:dataset_stats}. Further details about our difficulty annotation process are provided in Appendix \ref{app:difficulty}.

\subsection{Task Format}
The task format also serves as an important factor influencing reasoning performance \cite{Ribeiro2018SemanticallyEA,zong2024comparisonqaevaluatingfactualityrobustness}. We conducted experiments separately under two task formats\footnote{For the visual dataset, we evaluated only in the MCQ format for feasibility.}: multiple-choice questions (MCQ) and free-text generation (FTG), aiming to achieve a more comprehensive perspective in our exploration.

\begin{table}[t]
\centering
\small
(a) Modality (Task Format = MCQ)

\vspace{0.05cm}

\begin{tabular}{lccc}
\toprule
\multicolumn{1}{c}{} & \multicolumn{3}{c}{\textbf{Modality}} \\ \cmidrule(lr){2-4} 
\multicolumn{1}{c}{\multirow{-2}{*}{\textbf{Pipeline}}} & Textual & Visual & Symbolic \\ \midrule
Induction & 55.70 & 38.88 & 28.58 \\
Automatic & 58.05 & 51.52 & 34.99 \\
Abduction+Deduction & \textbf{59.13} & \textbf{53.93} & \textbf{37.69} \\ \midrule
System 2 Advantage & \cellcolor[HTML]{FFFFFF}{\color[HTML]{50BF50}+6.16\%} & \cellcolor[HTML]{FFFFFF}{\color[HTML]{50BF50}+38.73\%} & \cellcolor[HTML]{FFFFFF}{\color[HTML]{50BF50}+31.86\%} \\ \bottomrule
\vspace{0.01cm}
\end{tabular}

\setlength{\tabcolsep}{5.6pt}
(b) Difficulty (Task Format = MCQ)

\vspace{0.05cm}

\begin{tabular}{lccc}
\toprule
\multicolumn{1}{c}{} & \multicolumn{3}{c}{\textbf{Difficulty}} \\ \cmidrule(lr){2-4} 
\multicolumn{1}{c}{\multirow{-2}{*}{\textbf{Pipeline}}} & Easy & Medium & Hard \\ \midrule
Induction & 51.48 & 41.93 & 31.48 \\
Automatic & 58.12 & 48.23 & 40.02 \\
Abduction+Deduction & \textbf{59.68} & \textbf{49.76} & \textbf{43.20} \\ \midrule
System 2 Advantage & \multicolumn{1}{r}{\cellcolor[HTML]{FFFFFF}{\color[HTML]{50BF50}+15.92\%}} & \multicolumn{1}{r}{\cellcolor[HTML]{FFFFFF}{\color[HTML]{50BF50}+18.68\%}} & \multicolumn{1}{r}{\cellcolor[HTML]{FFFFFF}{\color[HTML]{50BF50}+37.20\%}} \\ \bottomrule
\vspace{0.01cm}
\end{tabular}
\setlength{\tabcolsep}{2.4pt}
(c) Task Format

\vspace{0.05cm}

\begin{tabular}{lccc>{\columncolor[HTML]{ffe1e1}}c}
\toprule
\multicolumn{1}{c}{} & \multicolumn{2}{c}{Textual} & \multicolumn{2}{c}{Symbolic} \\ \cmidrule(lr){2-3} \cmidrule(lr){4-5} 
\multicolumn{1}{c}{\multirow{-2}{*}{\textbf{Pipeline}}} & MCQ & FTG & MCQ & FTG \\ \midrule
Induction & 55.70 & 23.36 & 28.58 & \textbf{19.18} \\
Automatic & 58.05 & 24.89 & 34.99 & 8.67 \\
Abduction+Deduction & \textbf{59.13} & \textbf{24.93} & \textbf{37.69} & 11.33 \\ \midrule
System 2 Advantage & {\color[HTML]{50BF50}+6.16\%} & {\color[HTML]{50BF50}+6.74\%} & {\color[HTML]{50BF50}+31.86\%} & {\color[HTML]{FF0000} -40.93\%} \\ \bottomrule
\end{tabular}

\caption{Comparative dynamics of different logical inference pipelines in our evaluation environment, controlled by \textit{modality}, \textit{difficulty}, and \textit{task format}. Performances (in Accuracy \%) are averaged across all LLMs. "System 2 Advantage" denotes the relative improvements of abduction + deduction pipeline over direct induction.}
\label{tab:dynamics_modality}
\vspace{-0.3cm}
\end{table}

\section{Main Experiment Results and Analysis}
\label{sec:main}
We evaluated 10 modern LLMs / MLLMs (details provided in Appendix \ref{app:llm}) within our exploration environment. The experimental results are presented in Figure \ref{fig:main_result}. Across the entire environment, the tested LLMs achieved an overall average performance of only \textbf{35.4\%}, demonstrating that our datasets effectively stress-test the real reasoning abilities of LLMs rather than simply retrieving from memorization. Furthermore, the substantial performance gaps across difficulty levels validate the effectiveness of our difficulty annotations. Generally, the abduction + deduction pipeline outperforms direct induction, while automatic inference falls between the two pipelines in most scenarios.

To better illustrate the comparative dynamics between different logical inference pipelines, we present the consolidated results controlled by each dimension in Table \ref{tab:dynamics_modality}. From these results, we observe the key findings as follows:
\vspace{-0.1cm}
\paragraph{Findings 1: The comparative advantages of the System 2 logical inference pipeline are modality-dependent.} As shown in Table \ref{tab:dynamics_modality}(a), the abduction + deduction pipeline substantially outperforms direct induction in visual and symbolic tasks, with relative improvements of 38.73\% and 31.86\%, respectively. However, in textual tasks, direct induction achieves comparable performance, trailing behind by only 6.16\%.
\vspace{-0.1cm}
\paragraph{Findings 2: The comparative advantages of the System 2 logical inference pipeline are difficulty-dependent.} Based on Table \ref{tab:dynamics_modality}(b), the abduction + deduction pipeline outperforms direct induction by 37.20\% on hard questions, while the performance gap reduces to 18.68\% and 15.92\% on medium and easy questions, respectively.
\vspace{-0.1cm}
\paragraph{Findings 3: The System 2 logical inference pipeline falls short in free-text generation format when the task requires explicit rule execution.} Results from Table \ref{tab:dynamics_modality}(c) reveal a \textit{noteworthy inconsistency}: in textual tasks, the advantage of the System 2 pipeline remains the same across task formats. However, in symbolic tasks (i.e., RAVEN), the System 2 pipeline severely underperforms direct induction in the free-text generation format, which \textbf{sharply contrasts} with its advantage in the multiple-choice question format.

\paragraph{Interpretation of Findings 3:} To investigate the underlying mechanism leading to the limitation of System 2 logical inference in free-text generation, we conducted further analyses to decouple the performance of abduction and deduction (detailed in Appendix \ref{app:interpret}). We identified the following explanations for this task format sensitivity:

\begin{itemize}
\vspace{-0.2cm}

\item The precise generation of complex rules is challenging for most LLMs, as evidenced by the poor pattern inference accuracy compared to pattern execution (Table \ref{tab:adcompare}).
\vspace{-0.2cm}
\item Implicit pattern matching may be more effective in this case, as employed by direct induction. However, in the System 2 pipeline, lengthy rationales disrupt the well-structured few-shot patterns essential for in-context learning, thereby rendering implicit learning ineffective (Table \ref{tab:distance}).
\vspace{-0.2cm}
\item For multiple-choice questions, the System 2 pipeline can better infer patterns, as the answer space is reduced to a few candidates. It may also occasionally leverage reasoning shortcuts to improve performance \cite{geirhos2020shortcut,zong2024comparisonqaevaluatingfactualityrobustness} — an advantage that cannot be employed in direct induction.
\end{itemize}

As a result, the abduction + deduction pipeline tends to favor the MCQ format when addressing problems that require explicit rule execution, whereas, under the FTG format, direct induction demonstrates a surprising advantage.

\begin{figure}[t]
\begin{center}
\includegraphics[clip,width=220pt]{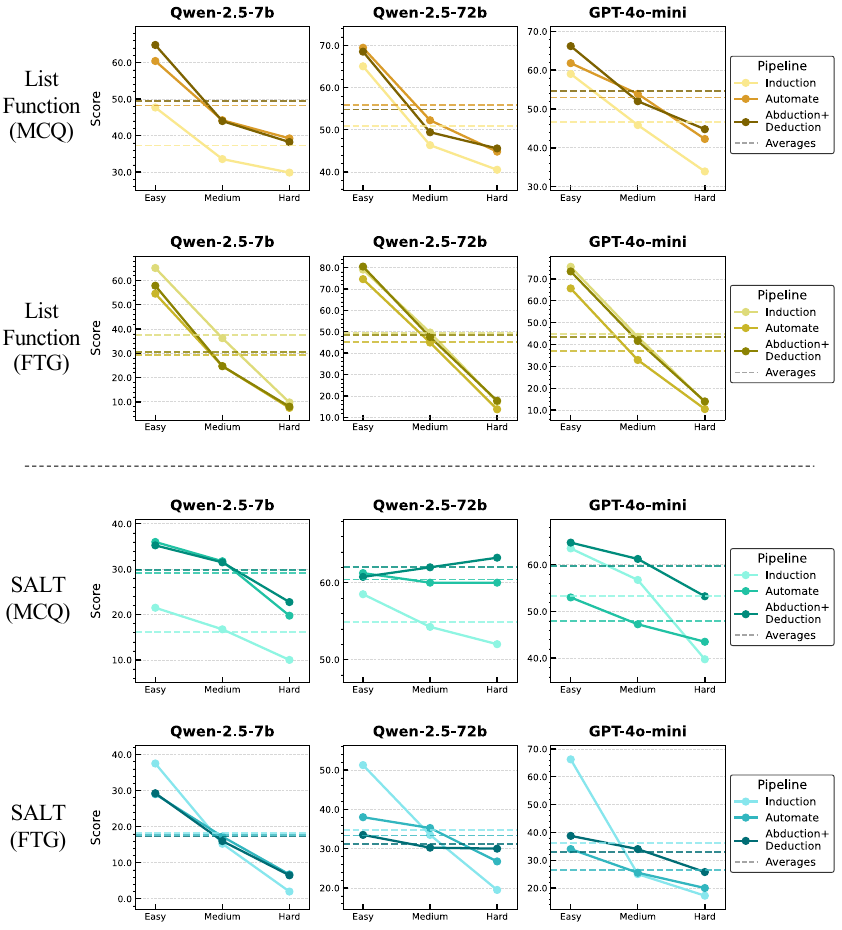}
\end{center}

\caption{LLM performances (in Accuracy \%) in List Function and SALT under different reasoning pipelines.}
\vspace{-0.3cm}
\label{fig:gen_result}
\end{figure}

\begin{table}[]
\setlength{\tabcolsep}{3.81pt}
\centering
\small
(a) List Function
\scriptsize

\begin{tabular}{lcccc>{\columncolor[HTML]{ffe1e1}}c}
\toprule
\multicolumn{1}{c}{} & \multicolumn{3}{c}{\textbf{Difficulty}} & \multicolumn{2}{c}{\textbf{Task Format}} \\ \cmidrule(lr){2-4} \cmidrule(lr){5-6}
\multicolumn{1}{c}{\multirow{-2}{*}{\textbf{Pipeline}}} & Easy & Medium & Hard & MCQ & FTG \\ \midrule
Induction & 65.26 & 42.53 & 24.18 & 44.96 & \textbf{43.92} \\
Automatic & 64.42 & 42.16 & 26.35 & 52.35 & 37.09 \\
Abduction+Deduction & \textbf{68.55} & \textbf{43.21} & \textbf{28.06} & \textbf{52.93} & 40.85 \\ \midrule
System 2 Advantage & {\color[HTML]{50BF50}+5.04\%} & {\color[HTML]{50BF50} +1.60\%} & {\color[HTML]{50BF50}+16.06\%} & {\color[HTML]{50BF50}+17.73\%} & {\color[HTML]{FF0000} -6.98\%} \\ \bottomrule
\vspace{-0.05cm}
\end{tabular}
\setlength{\tabcolsep}{3.5pt}
\small
(b) SALT
\scriptsize

\begin{tabular}{lcccc>{\columncolor[HTML]{ffe1e1}}c}
\toprule
\multicolumn{1}{c}{} & \multicolumn{3}{c}{\textbf{Difficulty}} & \multicolumn{2}{c}{\textbf{Task Format}} \\ \cmidrule(lr){2-4} \cmidrule(lr){5-6}
\multicolumn{1}{c}{\multirow{-2}{*}{\textbf{Pipeline}}} & Easy & Medium & Hard & MCQ & FTG \\ \midrule
Induction & \textbf{49.75} & 33.58 & 23.42 & 41.44 & \textbf{29.72} \\
Automatic & 41.88 & 36.17 & 29.46 & 45.83 & 25.83 \\
Abduction+Deduction & 43.71 & \textbf{39.17} & \textbf{33.58} & \textbf{50.53} & 27.11 \\ \midrule
System 2 Advantage & {\color[HTML]{FF0000}-12.14\%} & {\color[HTML]{50BF50} +16.63\%} & {\color[HTML]{50BF50}+43.42\%} & {\color[HTML]{50BF50}+21.92\%} & {\color[HTML]{FF0000} -8.79\%} \\ \bottomrule
\vspace{0.01cm}

\\\midrule
\multicolumn{1}{c}{\textbf{Average}}& {\color[HTML]{FF0000}-3.55\%} & {\color[HTML]{50BF50} +9.11\%} & {\color[HTML]{50BF50}+29.74\%} & {\color[HTML]{50BF50}+19.83\%} & {\color[HTML]{FF0000} -7.88\%} \\ \bottomrule
\end{tabular}

\caption{Comparative dynamics of different logical inference pipelines in List Function and SALT. Performances (in Accuracy \%) are averaged across all LLMs.}
\vspace{-0.3cm}
\label{tab:gen_dynamics}
\end{table}

\section{Generalization Experiment}
\label{sec:generalization}
To further assess the generalizability of our findings, we extend the scope from analogical reasoning to general in-context learning tasks. Specifically, we formally define our target task scope using the following constraints: 1) The task requires generating output \(y\) from input \(x\), based on \(n\)-shot demonstrations \(D = [(x_1, y_1), \dots, (x_n, y_n)]\). 2) The input-output function \(y = f(x)\) can be explicitly defined. We conduct generalization experiments on two in-context learning datasets, both of which require explicit rule execution. 

\vspace{0.2cm}
\noindent{\textbf{List Function}} \cite{rule2020child} takes lists of integers as input and maps them to output lists using 250 predefined transition functions. In this task, LLMs must infer the underlying function from provided demonstrations (input-output pairs) and apply it to new input lists. The difficulty of the task is determined by the complexity of the transition functions.

\vspace{0.2cm}
\noindent{\textbf{SALT}} (\textbf{S}yntax-aware \textbf{A}rtificial \textbf{L}anguage \textbf{T}rans- lation) is a machine translation benchmark that we developed to address key limitations in existing datasets. Unlike benchmarks such as SCAN \cite{higgins2018scanlearninghierarchicalcompositional} and Kalamang \cite{tanzer2024benchmarklearningtranslatenew}, SALT introduces diverse syntactic shifts (e.g., inversion of semantic unit order) while rigorously mitigating data leakage—a common issue in low-resource machine translation benchmarks. The task difficulty is determined by the complexity of the syntactic structures, enabling fine-grained evaluation of model performance across varying levels of linguistic challenge. Details of the SALT dataset are provided in Appendix \ref{app:salt}.

The results of the generalization experiments are illustrated in Figure \ref{fig:gen_result}, with the consolidated findings presented in Table \ref{tab:gen_dynamics}. Across both datasets,   we observed patterns similar to those in our evaluation environment in analogy: The advantage of the System 2 logical inference pipeline increases substantially as task difficulty rises. While the two pipelines exhibit contrasting task preferences between the MCQ and FTG format. Consequently, we demonstrate that \textbf{our findings are generalizable to broader in-context learning tasks} where the input-output function is explicitly defined.

\section{Scaling-up System 2 Logical Inference}
\label{sec:scaling}
Beyond the basic processes of abductive hypothesis generation and deductive execution (which form the core of our System 2 pipeline), more sophisticated logical inference strategies can be employed to tackle complex tasks and further enhance System 2 reasoning. We introduce two inference methodologies in philosophy and connect them to the logical inference pipelines of LLMs.

\subsection{Liptonian and Holmesian Inference}

\vspace{0.1cm}
\noindent{\textbf{Liptonian Inference}} \cite{Lipton2000-LIPITT-8} provides a widely recognized modern account of IBE (Inference to the Best Explanation). It characterizes the process of selecting the most explanatory hypothesis from a set of candidates based on its capacity to best account for the observed evidence.  In LLM reasoning, this corresponds to the parallel sampling of multiple hypotheses, followed by hypothesis selection as a precursor to the final deductive execution. In our experiment, we evaluated the effectiveness of \textbf{hypothesis selection} across sampling sizes ranging from 1 to 10.

\vspace{0.1cm}  
\noindent{\textbf{Holmesian Inference}} \cite{Bird2005-BIRAKA} provides an alternative model to Liptonian, emphasizing hypothesis verification rather than selection. Inspired by Sherlock Holmes’s famous dictum, it involves systematically eliminating all but one hypothesis to ensure that the remaining one is necessarily true. In LLM reasoning, this can be simulated through iterative verification and refinement (regeneration) of hypotheses, where candidate outputs are repeatedly evaluated and improved. In our experiment, we investigated \textbf{hypothesis verification and refinement} across iteration rounds up to 5.
\begin{figure}[t]
\begin{center}
\includegraphics[clip,width=210pt]{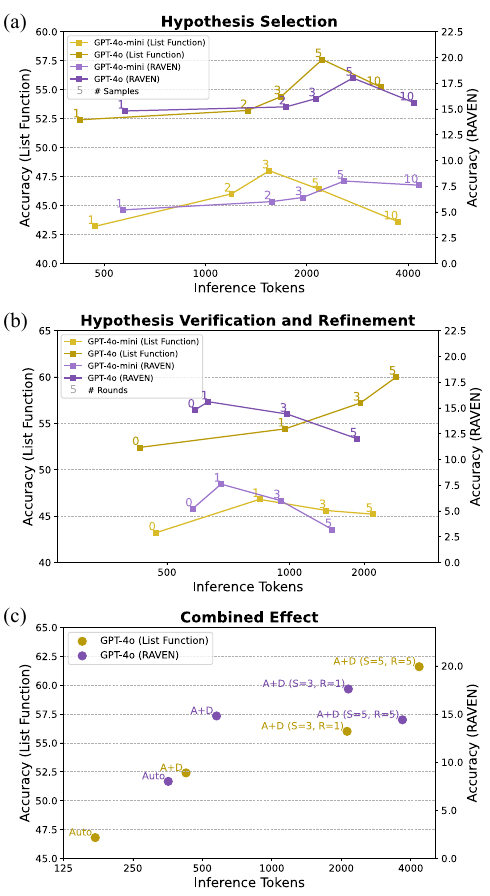}
\end{center}
\vspace{-0.3cm}
\caption{Effect of hypothesis selection, verification and refinement on LLM performances (in Accuracy \%).}
\vspace{-0.3cm}
\label{fig:scale_result}
\end{figure}
\subsection{Scaling Performances}
The experimental results of hypothesis selection, verification and refinement are presented in Figure \ref{fig:scale_result} (a) and \ref{fig:scale_result} (b). In hypothesis selection, we observe clear improvements in sampling sizes from 1 to 5. However, the performance starts to decrease when the sampling size increases to 10, as the diversity of the sampled hypotheses begins to saturate, and the selection process also becomes less effective with a longer context. In terms of hypothesis verification and refinement, the saturation of improvements was reached after one round of verification, except for GPT-4o in the List Function, where positive improvements were observed in every additional round of verification. This interesting inconsistency can be explained as follows: \textbf{1) Stronger LLMs lead to better verification quality.} Compared to the consistent improvements observed in GPT-4o, GPT-4o-mini did not exhibit similar enhancements, as its ability to detect incorrect hypotheses is also weaker.
\textbf{2) Well-formed hypothesis formats make refinements easier.} The improvement seen in the List-Function dataset (where hypotheses are written in Python code) does not hold for the RAVEN dataset (where hypotheses are presented in free text). A better hypothesis format may also enhance the effectiveness of proofreading or maintaining the validity of existing hypotheses.

Figure \ref{fig:scale_result} (c) illustrates the combined effect of the two scaling strategies. In both datasets, GPT-4o demonstrates considerable performance improvements as the number of inference tokens increases. For instance, performance of GPT-4o in the List-Function dataset improved from \textbf{46.8\%} to \textbf{61.6\%}, consuming \textbf{25×} more inference tokens compared to automatic inference. \textbf{This underscores the potential of scaling up LLM reasoning through System 2 logical inference pipelines.}

\setlength{\tabcolsep}{3.2pt}

\subsection{Discussions on Large Reasoning Models}
Recent advances in large reasoning\,models\,(LRMs), such as o1 \cite{openai2024o1preview} and Deepseek-R1 \cite{deepseekai2025deepseekr1incentivizingreasoningcapability}, have demonstrated impressive performance in mathematical and code reasoning tasks. LRMs emerge strong self-reflec- tion abilities during their reinforcement learning stage, driven by rule-based rewards. From\,our\,explor- ation, LRMs exhibit two noteworthy characteristics within our task domain (in-context learning with explicit input/output functions): 1) \textbf{LRMs emulate an "iterative holmesian inference"} by engaging in repeated cycles of hypothesis generation and verification. 2) \textbf{The number of inference tokens (rounds of iterative hypothesis generation) increases substantially as task difficulty rises}.

Nevertheless, can short-CoT LLMs achieve comparable performance by scaling up System 2 logical inference? To answer this question, we conducted experiments on Deepseek-V3 \cite{deepseekai2024deepseekv3technicalreport}, employing adaptive logical inference scaling under \textit{low} and \textit{high} computational consumptions (details in Appendix \ref{app:scale}), where the model autonomously determined the number of iteration within a set limit. As illustrated in Table~\ref{tab:lrm}, under high consumptions, Deepseek-V3 \textbf{demonstrates a similar inference scaling effect in difficulty and achieves comparable performance to LRMs}.

\begin{table}[]
\centering
\scriptsize
\begin{tabular}{lcccc}
\toprule
\multicolumn{1}{c}{\multirow{2}{*}{\textbf{Model}}} & \multicolumn{3}{c}{\textbf{Inference Tokens (\# Rounds)}} & \multirow{2}{*}{\textbf{Accuracy}} \\ \cmidrule(lr){2-4}
\multicolumn{1}{c}{} & \textbf{Easy} & \textbf{Medium} & \textbf{Hard} &  \\ \midrule
Deepseek-R1 & 2174.5 (3.9) & 3353.1 (5.0) & 5935.9 (6.5) & 69.2 \\
o1-mini & 1345.5 (2.6) & 2229.8 (3.2) & 4188.0 (3.5) & 69.6 \\
o1 & 1949.1 (2.7) & 3233.0 (3.3) & 6995.7 (5.5) & \textbf{77.2} \\
o3-mini & 1184.3 (2.5) & 2126.3 (3.0) & 5328.7 (6.2) & \underline{76.8} \\ \midrule
\vspace{-0.15cm}

\\ \midrule
Deepseek-V3 & 989.0 & 1261.1 & 1260.9 & 57.2 \\
 +Sys2 Scaling (low) & 1758.0 (2.4) & 2124.4 (2.5) & 2618.9 (2.7) & 65.2 \\
 +Sys2 Scaling (high) & 2356.8 (2.7) & 2985.3 (2.9) & 4308.0 (3.8) & 69.6 \\\bottomrule
\end{tabular}
\caption{Performance of LRMs and LLMs with adaptive logical inference scaling on the List Function dataset.}
\vspace{-0.4cm}
\label{tab:lrm}
\end{table}

\section{Related Work}
\subsection{Logical Inference in Language Models}
\paragraph{Abductive Inference}
In the era of pre-trained language models, $\alpha$\textbf{-NLI} \citep{bhagavatula2020abductivecommonsensereasoning} introduced abductive reasoning to commonsense reasoning, where plausible explanations are inferred from observations. Subsequent works proposed various techniques to enhance this capability \citep{qin2021futureunsupervisedbackpropbaseddecoding, kadiķis2022embarrassinglysimpleperformanceprediction, chan2023selfconsistentnarrativepromptsabductive}, including extensions to uncommon scenarios focusing on rare but logical explanations \citep{zhao2024uncommonsensereasoningabductivereasoning}. Unlike real-world data in commonsense reasoning, benchmarks like \textbf{ProofWriter} \citep{tafjord2021proofwritergeneratingimplicationsproofs} evaluate formal abductive reasoning within semi-structured texts with explicit logical relationships. Recent studies have explored LLMs in more challenging open-world reasoning contexts \citep{zhong2023chatablabductivelearningnatural, del2023truedetectivedeepabductive, thagard2024chatgptmakeexplanatoryinferences}. Beyond natural language inference, abductive reasoning has also been examined in graph-based modalities for commonsense and event knowledge \citep{du-etal-2021-learning, bai2024advancingabductivereasoningknowledge}.

\paragraph{Deductive and Inductive Inference}
Deductive inference is extensively studied in transformers \citep{clark2020transformerssoftreasonerslanguage,han2024folionaturallanguagereasoning,zheng-etal-2025-enhancing-transformers} with natural language rule-based reasoning tasks. \citet{saparov2023testinggeneraldeductivereasoning} evaluate LLMs' deductive reasoning in out-of-distribution settings, emphasizing challenges with longer proofs and complex logic. Inductive inference is explored through datasets like \textbf{EntailmentBank} \citep{dalvi2022explaininganswersentailmenttrees}, where models construct step-by-step entailment trees to explain answers. Meanwhile, recent studies have demonstrated emergent inductive reasoning abilities in LLMs \citep{zheng2025cursecotlimitationschainofthought,li-etal-2025-patterns,fan2025legalruleinductiongeneralizable}.

\subsection{Analogical Reasoning} 
The study of analogical reasoning in AI has progressed from early symbolic systems, such as the \textbf{Structure-Mapping Engine} \citep{FALKENHAINER19891}, which used hand-crafted representations, to models like the \textbf{Latent Relation Mapping Engine} \citep{Turney_2008}, which integrated symbolic rules with statistical learning. The neural era introduced word embeddings for analogy evaluation \citep{mikolov2013distributedrepresentationswordsphrases}, emphasizing local semantic patterns. With LLMs, \citet{webb2023emergentanalogicalreasoninglarge} demonstrated emergent analogical reasoning, but challenges remain. \textbf{AnaloBench} \citep{ye2024analobenchbenchmarkingidentificationabstract} shows minimal scaling gains for long-context analogies, while \textbf{ANALOGICAL} \citep{wijesiriwardene2023analogicalnovelbenchmark} highlights struggles with complex metaphors. Story-level benchmarks like \textbf{StoryAnalogy} \citep{jiayang2023storyanalogyderivingstorylevelanalogies} and \textbf{ARN} \citep{sourati2024arnanalogicalreasoningnarratives} reveal difficulties in cross-domain narrative mapping without explicit prompts.
\vspace{-0.1cm}
\section{Conclusion}
\vspace{-0.1cm}
This paper systematically dissects the interplay of inductive (System 1) and abductive/deductive (System 2) logical inference within LLMs. 
We establish that while System 2 pipelines generally yield superior performance—particularly in visual/symbolic modalities and with increasing task difficulty—System 1 remains competitive for textual tasks and, crucially, can outperform System 2 in free-text rule-execution scenarios. 
These nuanced dynamics extend to broader ICL tasks involving explicit input-output functions. 
Furthermore, we demonstrate that strategically scaling System 2 through methods like hypothesis selection and iterative refinement largely enhances reasoning capabilities, enabling standard LLMs to approach the performance of specialized reasoning models. 
Ultimately, this study provides a foundational understanding and actionable guidelines for optimizing LLM reasoning by tailoring logical inference strategies to specific task characteristics.

\section*{Limitations}
While our extensive experiments and analyses yield rich findings, our exploration is limited to reasoning frameworks for static LLMs. Future research could build on this work by focusing on the tuning stage of LLMs, aiming to develop systems that dynamically balance different types of logical inference. For example, a system capable of automatically identifying the nature of a question and determining whether to apply System 1 or System 2 reasoning could not only maintain or enhance performance but also improve efficiency. Such adaptive reasoning closely mirrors the way humans naturally approach problem-solving.

\section*{Ethics Statement}

This work aims to advance the understanding of logical inference in LLMs through systematic experimentation and analysis. All LLMs used in this study are publicly available. We strictly prohibit harmful content in the selection, curation, and annotation process of our datasets, ensuring they are free from sensitive or biased material. Our work is conducted with a focus on advancing understanding while adhering to ethical research practices.

\section*{Acknowledgements}

We thank all the anonymous reviewers and meta reviewers for their valuable comments. The authors of this paper were supported by the ITSP Platform Research Project (ITS/189/23FP) from ITC of Hong Kong, SAR, China, and the AoE (AoE/E-601/24-N), the RIF (R6021-20) and the GRF (16205322) from RGC of Hong Kong, SAR, China. We also thank the support from NVIDIA AI Technology Center (NVAITC).

\bibliography{main}

\begin{thebibliography}{80}
\providecommand{\natexlab}[1]{#1}

\bibitem[{Agrawal et~al.(2024)Agrawal, Antoniak, Hanna, Bout, Chaplot, Chudnovsky, Costa, Monicault, Garg, Gervet, Ghosh, Héliou, Jacob, Jiang, Khandelwal, Lacroix, Lample, Casas, Lavril, Scao, Lo, Marshall, Martin, Mensch, Muddireddy, Nemychnikova, Pellat, Platen, Raghuraman, Rozière, Sablayrolles, Saulnier, Sauvestre, Shang, Soletskyi, Stewart, Stock, Studnia, Subramanian, Vaze, Wang, and Yang}]{agrawal2024pixtral12b}
Pravesh Agrawal, Szymon Antoniak, Emma~Bou Hanna, Baptiste Bout, Devendra Chaplot, Jessica Chudnovsky, Diogo Costa, Baudouin~De Monicault, Saurabh Garg, Theophile Gervet, Soham Ghosh, Amélie Héliou, Paul Jacob, Albert~Q. Jiang, Kartik Khandelwal, Timothée Lacroix, Guillaume Lample, Diego~Las Casas, Thibaut Lavril, Teven~Le Scao, Andy Lo, William Marshall, Louis Martin, Arthur Mensch, Pavankumar Muddireddy, Valera Nemychnikova, Marie Pellat, Patrick~Von Platen, Nikhil Raghuraman, Baptiste Rozière, Alexandre Sablayrolles, Lucile Saulnier, Romain Sauvestre, Wendy Shang, Roman Soletskyi, Lawrence Stewart, Pierre Stock, Joachim Studnia, Sandeep Subramanian, Sagar Vaze, Thomas Wang, and Sophia Yang. 2024.
\newblock \href {https://arxiv.org/abs/2410.07073} {Pixtral 12b}.
\newblock \emph{Preprint}, arXiv:2410.07073.

\bibitem[{AI(2024)}]{meta2024llama31}
Meta AI. 2024.
\newblock \href {https://ai.meta.com/blog/meta-llama-3-1/} {Introducing {Llama} 3.1: Our most capable models to date}.

\bibitem[{Bai et~al.(2024)Bai, Wang, Zheng, Guo, Liu, and Song}]{bai2024advancingabductivereasoningknowledge}
Jiaxin Bai, Yicheng Wang, Tianshi Zheng, Yue Guo, Xin Liu, and Yangqiu Song. 2024.
\newblock \href {https://arxiv.org/abs/2312.15643} {Advancing abductive reasoning in knowledge graphs through complex logical hypothesis generation}.
\newblock \emph{Preprint}, arXiv:2312.15643.

\bibitem[{Bhagavatula et~al.(2020)Bhagavatula, Bras, Malaviya, Sakaguchi, Holtzman, Rashkin, Downey, tau Yih, and Choi}]{bhagavatula2020abductivecommonsensereasoning}
Chandra Bhagavatula, Ronan~Le Bras, Chaitanya Malaviya, Keisuke Sakaguchi, Ari Holtzman, Hannah Rashkin, Doug Downey, Scott~Wen tau Yih, and Yejin Choi. 2020.
\newblock \href {https://arxiv.org/abs/1908.05739} {Abductive commonsense reasoning}.
\newblock \emph{Preprint}, arXiv:1908.05739.

\bibitem[{Bird(2005)}]{Bird2005-BIRAKA}
Alexander Bird. 2005.
\newblock Abductive knowledge and holmesian inference.
\newblock In Tamar~Szab\'o Gendler and John Hawthorne, editors, \emph{Oxford Studies in Epistemology}, pages 1--31. Oxford University Press.

\bibitem[{Bitton et~al.(2022)Bitton, Yosef, Strugo, Shahaf, Schwartz, and Stanovsky}]{bitton2022vasrvisualanalogiessituation}
Yonatan Bitton, Ron Yosef, Eli Strugo, Dafna Shahaf, Roy Schwartz, and Gabriel Stanovsky. 2022.
\newblock \href {https://arxiv.org/abs/2212.04542} {Vasr: Visual analogies of situation recognition}.
\newblock \emph{Preprint}, arXiv:2212.04542.

\bibitem[{Bojanowski et~al.(2017)Bojanowski, Grave, Joulin, and Mikolov}]{bojanowski2017enriching}
Piotr Bojanowski, Edouard Grave, Armand Joulin, and Tomas Mikolov. 2017.
\newblock Enriching word vectors with subword information.
\newblock \emph{Transactions of the Association for Computational Linguistics}, 5:135--146.

\bibitem[{Chan et~al.(2023)Chan, Liu, Chan, Cheng, Song, Wong, and See}]{chan2023selfconsistentnarrativepromptsabductive}
Chunkit Chan, Xin Liu, Tsz~Ho Chan, Jiayang Cheng, Yangqiu Song, Ginny Wong, and Simon See. 2023.
\newblock \href {https://arxiv.org/abs/2309.08303} {Self-consistent narrative prompts on abductive natural language inference}.
\newblock \emph{Preprint}, arXiv:2309.08303.

\bibitem[{Chen et~al.(2022)Chen, Xu, Fu, Shi, Li, Zhang, Sun, Li, Xiao, and Zhou}]{Chen_2022}
Jiangjie Chen, Rui Xu, Ziquan Fu, Wei Shi, Zhongqiao Li, Xinbo Zhang, Changzhi Sun, Lei Li, Yanghua Xiao, and Hao Zhou. 2022.
\newblock \href {https://doi.org/10.18653/v1/2022.findings-acl.311} {E-kar: A benchmark for rationalizing natural language analogical reasoning}.
\newblock In \emph{Findings of the Association for Computational Linguistics: ACL 2022}, page 3941–3955. Association for Computational Linguistics.

\bibitem[{Chollet(2019)}]{chollet2019measure}
Fran{\c{c}}ois Chollet. 2019.
\newblock On the measure of intelligence.
\newblock \emph{arXiv preprint arXiv:1911.01547}.

\bibitem[{Clark et~al.(2020)Clark, Tafjord, and Richardson}]{clark2020transformerssoftreasonerslanguage}
Peter Clark, Oyvind Tafjord, and Kyle Richardson. 2020.
\newblock \href {https://arxiv.org/abs/2002.05867} {Transformers as soft reasoners over language}.
\newblock \emph{Preprint}, arXiv:2002.05867.

\bibitem[{Cohen(1982)}]{2af2fe0f-5f65-3dca-bfa3-cd96adbc1502}
L.~Jonathan Cohen. 1982.
\newblock \href {http://www.jstor.org/stable/27902756} {Intuition, induction, and the middle way}.
\newblock \emph{The Monist}, 65(3):287--301.

\bibitem[{Copi and Cohen(1990)}]{copi1990introduction}
I.M. Copi and C.~Cohen. 1990.
\newblock \href {https://books.google.com.hk/books?id=UnIbAQAAMAAJ} {\emph{Introduction to Logic}}.
\newblock Maxwell Macmillan international editions. Macmillan.

\bibitem[{Dalvi et~al.(2022)Dalvi, Jansen, Tafjord, Xie, Smith, Pipatanangkura, and Clark}]{dalvi2022explaininganswersentailmenttrees}
Bhavana Dalvi, Peter Jansen, Oyvind Tafjord, Zhengnan Xie, Hannah Smith, Leighanna Pipatanangkura, and Peter Clark. 2022.
\newblock \href {https://arxiv.org/abs/2104.08661} {Explaining answers with entailment trees}.
\newblock \emph{Preprint}, arXiv:2104.08661.

\bibitem[{DeepMind(2024)}]{google2024gemini2}
Google DeepMind. 2024.
\newblock \href {https://blog.google/technology/google-deepmind/google-gemini-ai-update-december-2024/} {Google introduces {Gemini} 2.0: A new {AI} model for the agentic era}.

\bibitem[{DeepSeek-AI et~al.(2025)DeepSeek-AI, Guo, Yang, Zhang, Song, Zhang, Xu, Zhu, Ma, Wang, Bi, Zhang, Yu, Wu, Wu, Gou, Shao, Li, Gao, Liu, Xue, Wang, Wu, Feng, Lu, Zhao, Deng, Zhang, Ruan, Dai, Chen, Ji, Li, Lin, Dai, Luo, Hao, Chen, Li, Zhang, Bao, Xu, Wang, Ding, Xin, Gao, Qu, Li, Guo, Li, Wang, Chen, Yuan, Qiu, Li, Cai, Ni, Liang, Chen, Dong, Hu, Gao, Guan, Huang, Yu, Wang, Zhang, Zhao, Wang, Zhang, Xu, Xia, Zhang, Zhang, Tang, Li, Wang, Li, Tian, Huang, Zhang, Wang, Chen, Du, Ge, Zhang, Pan, Wang, Chen, Jin, Chen, Lu, Zhou, Chen, Ye, Wang, Yu, Zhou, Pan, Li, Zhou, Wu, Ye, Yun, Pei, Sun, Wang, Zeng, Zhao, Liu, Liang, Gao, Yu, Zhang, Xiao, An, Liu, Wang, Chen, Nie, Cheng, Liu, Xie, Liu, Yang, Li, Su, Lin, Li, Jin, Shen, Chen, Sun, Wang, Song, Zhou, Wang, Shan, Li, Wang, Wei, Zhang, Xu, Li, Zhao, Sun, Wang, Yu, Zhang, Shi, Xiong, He, Piao, Wang, Tan, Ma, Liu, Guo, Ou, Wang, Gong, Zou, He, Xiong, Luo, You, Liu, Zhou, Zhu, Xu, Huang, Li, Zheng, Zhu, Ma, Tang, Zha, Yan, Ren, Ren, Sha, Fu, Xu, Xie, Zhang,
  Hao, Ma, Yan, Wu, Gu, Zhu, Liu, Li, Xie, Song, Pan, Huang, Xu, Zhang, and Zhang}]{deepseekai2025deepseekr1incentivizingreasoningcapability}
DeepSeek-AI, Daya Guo, Dejian Yang, Haowei Zhang, Junxiao Song, Ruoyu Zhang, Runxin Xu, Qihao Zhu, Shirong Ma, Peiyi Wang, Xiao Bi, Xiaokang Zhang, Xingkai Yu, Yu~Wu, Z.~F. Wu, Zhibin Gou, Zhihong Shao, Zhuoshu Li, Ziyi Gao, Aixin Liu, Bing Xue, Bingxuan Wang, Bochao Wu, Bei Feng, Chengda Lu, Chenggang Zhao, Chengqi Deng, Chenyu Zhang, Chong Ruan, Damai Dai, Deli Chen, Dongjie Ji, Erhang Li, Fangyun Lin, Fucong Dai, Fuli Luo, Guangbo Hao, Guanting Chen, Guowei Li, H.~Zhang, Han Bao, Hanwei Xu, Haocheng Wang, Honghui Ding, Huajian Xin, Huazuo Gao, Hui Qu, Hui Li, Jianzhong Guo, Jiashi Li, Jiawei Wang, Jingchang Chen, Jingyang Yuan, Junjie Qiu, Junlong Li, J.~L. Cai, Jiaqi Ni, Jian Liang, Jin Chen, Kai Dong, Kai Hu, Kaige Gao, Kang Guan, Kexin Huang, Kuai Yu, Lean Wang, Lecong Zhang, Liang Zhao, Litong Wang, Liyue Zhang, Lei Xu, Leyi Xia, Mingchuan Zhang, Minghua Zhang, Minghui Tang, Meng Li, Miaojun Wang, Mingming Li, Ning Tian, Panpan Huang, Peng Zhang, Qiancheng Wang, Qinyu Chen, Qiushi Du, Ruiqi Ge, Ruisong
  Zhang, Ruizhe Pan, Runji Wang, R.~J. Chen, R.~L. Jin, Ruyi Chen, Shanghao Lu, Shangyan Zhou, Shanhuang Chen, Shengfeng Ye, Shiyu Wang, Shuiping Yu, Shunfeng Zhou, Shuting Pan, S.~S. Li, Shuang Zhou, Shaoqing Wu, Shengfeng Ye, Tao Yun, Tian Pei, Tianyu Sun, T.~Wang, Wangding Zeng, Wanjia Zhao, Wen Liu, Wenfeng Liang, Wenjun Gao, Wenqin Yu, Wentao Zhang, W.~L. Xiao, Wei An, Xiaodong Liu, Xiaohan Wang, Xiaokang Chen, Xiaotao Nie, Xin Cheng, Xin Liu, Xin Xie, Xingchao Liu, Xinyu Yang, Xinyuan Li, Xuecheng Su, Xuheng Lin, X.~Q. Li, Xiangyue Jin, Xiaojin Shen, Xiaosha Chen, Xiaowen Sun, Xiaoxiang Wang, Xinnan Song, Xinyi Zhou, Xianzu Wang, Xinxia Shan, Y.~K. Li, Y.~Q. Wang, Y.~X. Wei, Yang Zhang, Yanhong Xu, Yao Li, Yao Zhao, Yaofeng Sun, Yaohui Wang, Yi~Yu, Yichao Zhang, Yifan Shi, Yiliang Xiong, Ying He, Yishi Piao, Yisong Wang, Yixuan Tan, Yiyang Ma, Yiyuan Liu, Yongqiang Guo, Yuan Ou, Yuduan Wang, Yue Gong, Yuheng Zou, Yujia He, Yunfan Xiong, Yuxiang Luo, Yuxiang You, Yuxuan Liu, Yuyang Zhou, Y.~X. Zhu,
  Yanhong Xu, Yanping Huang, Yaohui Li, Yi~Zheng, Yuchen Zhu, Yunxian Ma, Ying Tang, Yukun Zha, Yuting Yan, Z.~Z. Ren, Zehui Ren, Zhangli Sha, Zhe Fu, Zhean Xu, Zhenda Xie, Zhengyan Zhang, Zhewen Hao, Zhicheng Ma, Zhigang Yan, Zhiyu Wu, Zihui Gu, Zijia Zhu, Zijun Liu, Zilin Li, Ziwei Xie, Ziyang Song, Zizheng Pan, Zhen Huang, Zhipeng Xu, Zhongyu Zhang, and Zhen Zhang. 2025.
\newblock \href {https://arxiv.org/abs/2501.12948} {Deepseek-r1: Incentivizing reasoning capability in llms via reinforcement learning}.
\newblock \emph{Preprint}, arXiv:2501.12948.

\bibitem[{DeepSeek-AI et~al.(2024)DeepSeek-AI, Liu, Feng, Xue, Wang, Wu, Lu, Zhao, Deng, Zhang, Ruan, Dai, Guo, Yang, Chen, Ji, Li, Lin, Dai, Luo, Hao, Chen, Li, Zhang, Bao, Xu, Wang, Zhang, Ding, Xin, Gao, Li, Qu, Cai, Liang, Guo, Ni, Li, Wang, Chen, Chen, Yuan, Qiu, Li, Song, Dong, Hu, Gao, Guan, Huang, Yu, Wang, Zhang, Xu, Xia, Zhao, Wang, Zhang, Li, Wang, Zhang, Zhang, Tang, Li, Tian, Huang, Wang, Zhang, Wang, Zhu, Chen, Du, Chen, Jin, Ge, Zhang, Pan, Wang, Xu, Zhang, Chen, Li, Lu, Zhou, Chen, Wu, Ye, Ye, Ma, Wang, Zhou, Yu, Zhou, Pan, Wang, Yun, Pei, Sun, Xiao, Zeng, Zhao, An, Liu, Liang, Gao, Yu, Zhang, Li, Jin, Wang, Bi, Liu, Wang, Shen, Chen, Zhang, Chen, Nie, Sun, Wang, Cheng, Liu, Xie, Liu, Yu, Song, Shan, Zhou, Yang, Li, Su, Lin, Li, Wang, Wei, Zhu, Zhang, Xu, Xu, Huang, Li, Zhao, Sun, Li, Wang, Yu, Zheng, Zhang, Shi, Xiong, He, Tang, Piao, Wang, Tan, Ma, Liu, Guo, Wu, Ou, Zhu, Wang, Gong, Zou, He, Zha, Xiong, Ma, Yan, Luo, You, Liu, Zhou, Wu, Ren, Ren, Sha, Fu, Xu, Huang, Zhang, Xie, Zhang, Hao,
  Gou, Ma, Yan, Shao, Xu, Wu, Zhang, Li, Gu, Zhu, Liu, Li, Xie, Song, Gao, and Pan}]{deepseekai2024deepseekv3technicalreport}
DeepSeek-AI, Aixin Liu, Bei Feng, Bing Xue, Bingxuan Wang, Bochao Wu, Chengda Lu, Chenggang Zhao, Chengqi Deng, Chenyu Zhang, Chong Ruan, Damai Dai, Daya Guo, Dejian Yang, Deli Chen, Dongjie Ji, Erhang Li, Fangyun Lin, Fucong Dai, Fuli Luo, Guangbo Hao, Guanting Chen, Guowei Li, H.~Zhang, Han Bao, Hanwei Xu, Haocheng Wang, Haowei Zhang, Honghui Ding, Huajian Xin, Huazuo Gao, Hui Li, Hui Qu, J.~L. Cai, Jian Liang, Jianzhong Guo, Jiaqi Ni, Jiashi Li, Jiawei Wang, Jin Chen, Jingchang Chen, Jingyang Yuan, Junjie Qiu, Junlong Li, Junxiao Song, Kai Dong, Kai Hu, Kaige Gao, Kang Guan, Kexin Huang, Kuai Yu, Lean Wang, Lecong Zhang, Lei Xu, Leyi Xia, Liang Zhao, Litong Wang, Liyue Zhang, Meng Li, Miaojun Wang, Mingchuan Zhang, Minghua Zhang, Minghui Tang, Mingming Li, Ning Tian, Panpan Huang, Peiyi Wang, Peng Zhang, Qiancheng Wang, Qihao Zhu, Qinyu Chen, Qiushi Du, R.~J. Chen, R.~L. Jin, Ruiqi Ge, Ruisong Zhang, Ruizhe Pan, Runji Wang, Runxin Xu, Ruoyu Zhang, Ruyi Chen, S.~S. Li, Shanghao Lu, Shangyan Zhou, Shanhuang
  Chen, Shaoqing Wu, Shengfeng Ye, Shengfeng Ye, Shirong Ma, Shiyu Wang, Shuang Zhou, Shuiping Yu, Shunfeng Zhou, Shuting Pan, T.~Wang, Tao Yun, Tian Pei, Tianyu Sun, W.~L. Xiao, Wangding Zeng, Wanjia Zhao, Wei An, Wen Liu, Wenfeng Liang, Wenjun Gao, Wenqin Yu, Wentao Zhang, X.~Q. Li, Xiangyue Jin, Xianzu Wang, Xiao Bi, Xiaodong Liu, Xiaohan Wang, Xiaojin Shen, Xiaokang Chen, Xiaokang Zhang, Xiaosha Chen, Xiaotao Nie, Xiaowen Sun, Xiaoxiang Wang, Xin Cheng, Xin Liu, Xin Xie, Xingchao Liu, Xingkai Yu, Xinnan Song, Xinxia Shan, Xinyi Zhou, Xinyu Yang, Xinyuan Li, Xuecheng Su, Xuheng Lin, Y.~K. Li, Y.~Q. Wang, Y.~X. Wei, Y.~X. Zhu, Yang Zhang, Yanhong Xu, Yanhong Xu, Yanping Huang, Yao Li, Yao Zhao, Yaofeng Sun, Yaohui Li, Yaohui Wang, Yi~Yu, Yi~Zheng, Yichao Zhang, Yifan Shi, Yiliang Xiong, Ying He, Ying Tang, Yishi Piao, Yisong Wang, Yixuan Tan, Yiyang Ma, Yiyuan Liu, Yongqiang Guo, Yu~Wu, Yuan Ou, Yuchen Zhu, Yuduan Wang, Yue Gong, Yuheng Zou, Yujia He, Yukun Zha, Yunfan Xiong, Yunxian Ma, Yuting Yan, Yuxiang
  Luo, Yuxiang You, Yuxuan Liu, Yuyang Zhou, Z.~F. Wu, Z.~Z. Ren, Zehui Ren, Zhangli Sha, Zhe Fu, Zhean Xu, Zhen Huang, Zhen Zhang, Zhenda Xie, Zhengyan Zhang, Zhewen Hao, Zhibin Gou, Zhicheng Ma, Zhigang Yan, Zhihong Shao, Zhipeng Xu, Zhiyu Wu, Zhongyu Zhang, Zhuoshu Li, Zihui Gu, Zijia Zhu, Zijun Liu, Zilin Li, Ziwei Xie, Ziyang Song, Ziyi Gao, and Zizheng Pan. 2024.
\newblock \href {https://arxiv.org/abs/2412.19437} {Deepseek-v3 technical report}.
\newblock \emph{Preprint}, arXiv:2412.19437.

\bibitem[{Del and Fishel(2023)}]{del2023truedetectivedeepabductive}
Maksym Del and Mark Fishel. 2023.
\newblock \href {https://arxiv.org/abs/2212.10114} {True detective: A deep abductive reasoning benchmark undoable for gpt-3 and challenging for gpt-4}.
\newblock \emph{Preprint}, arXiv:2212.10114.

\bibitem[{Devlin et~al.(2019)Devlin, Chang, Lee, and Toutanova}]{devlin2019bertpretrainingdeepbidirectional}
Jacob Devlin, Ming-Wei Chang, Kenton Lee, and Kristina Toutanova. 2019.
\newblock \href {https://arxiv.org/abs/1810.04805} {Bert: Pre-training of deep bidirectional transformers for language understanding}.
\newblock \emph{Preprint}, arXiv:1810.04805.

\bibitem[{Du et~al.(2021)Du, Ding, Liu, and Qin}]{du-etal-2021-learning}
Li~Du, Xiao Ding, Ting Liu, and Bing Qin. 2021.
\newblock \href {https://doi.org/10.18653/v1/2021.acl-long.403} {Learning event graph knowledge for abductive reasoning}.
\newblock In \emph{Proceedings of the 59th Annual Meeting of the Association for Computational Linguistics and the 11th International Joint Conference on Natural Language Processing (Volume 1: Long Papers)}, pages 5181--5190, Online. Association for Computational Linguistics.

\bibitem[{Falkenhainer et~al.(1989)Falkenhainer, Forbus, and Gentner}]{FALKENHAINER19891}
Brian Falkenhainer, Kenneth~D. Forbus, and Dedre Gentner. 1989.
\newblock \href {https://doi.org/10.1016/0004-3702(89)90077-5} {The structure-mapping engine: Algorithm and examples}.
\newblock \emph{Artificial Intelligence}, 41(1):1--63.

\bibitem[{Fan et~al.(2025)Fan, Zheng, Hu, Deng, Wang, Xu, Li, Li, Shen, and Song}]{fan2025legalruleinductiongeneralizable}
Wei Fan, Tianshi Zheng, Yiran Hu, Zheye Deng, Weiqi Wang, Baixuan Xu, Chunyang Li, Haoran Li, Weixing Shen, and Yangqiu Song. 2025.
\newblock \href {https://arxiv.org/abs/2505.14104} {Legal rule induction: Towards generalizable principle discovery from analogous judicial precedents}.
\newblock \emph{Preprint}, arXiv:2505.14104.

\bibitem[{Flach and Kakas(2000)}]{flach2000abduction}
Peter~A Flach and Antonis~C Kakas. 2000.
\newblock \emph{Abduction and Induction: Essays on their Relation and Integration}.
\newblock Springer Science.

\bibitem[{Frankfurt(1958)}]{frankfurt1958peirce}
Harry Frankfurt. 1958.
\newblock Peirce's notion of abduction.
\newblock \emph{Journal of Philosophy}, 55:593--596.

\bibitem[{Geirhos et~al.(2020)Geirhos, Jacobsen, Michaelis, Zemel, Brendel, Bethge, and Wichmann}]{geirhos2020shortcut}
Robert Geirhos, J{\"o}rn-Henrik Jacobsen, Claudio Michaelis, Richard Zemel, Wieland Brendel, Matthias Bethge, and Felix~A. Wichmann. 2020.
\newblock \href {https://doi.org/10.1038/s42256-020-00257-z} {Shortcut learning in deep neural networks}.
\newblock \emph{Nature Machine Intelligence}, 2:665--673.

\bibitem[{Gentner et~al.(2001)Gentner, Holyoak, and Kokinov}]{gentner2001analogical}
Dedre Gentner, Keith Holyoak, and Boicho Kokinov. 2001.
\newblock \emph{The Analogical Mind: Perspectives From Cognitive Science}.
\newblock MIT Press.

\bibitem[{Google(2024)}]{google2024gemini}
Google. 2024.
\newblock \href {https://blog.google/technology/ai/google-gemini-next-generation-model-february-2024/} {Introducing {Gemini} 1.5, google's next-generation {AI} model}.

\bibitem[{Han et~al.(2024)Han, Schoelkopf, Zhao, Qi, Riddell, Zhou, Coady, Peng, Qiao, Benson, Sun, Wardle-Solano, Szabo, Zubova, Burtell, Fan, Liu, Wong, Sailor, Ni, Nan, Kasai, Yu, Zhang, Fabbri, Kryscinski, Yavuz, Liu, Lin, Joty, Zhou, Xiong, Ying, Cohan, and Radev}]{han2024folionaturallanguagereasoning}
Simeng Han, Hailey Schoelkopf, Yilun Zhao, Zhenting Qi, Martin Riddell, Wenfei Zhou, James Coady, David Peng, Yujie Qiao, Luke Benson, Lucy Sun, Alex Wardle-Solano, Hannah Szabo, Ekaterina Zubova, Matthew Burtell, Jonathan Fan, Yixin Liu, Brian Wong, Malcolm Sailor, Ansong Ni, Linyong Nan, Jungo Kasai, Tao Yu, Rui Zhang, Alexander~R. Fabbri, Wojciech Kryscinski, Semih Yavuz, Ye~Liu, Xi~Victoria Lin, Shafiq Joty, Yingbo Zhou, Caiming Xiong, Rex Ying, Arman Cohan, and Dragomir Radev. 2024.
\newblock \href {https://arxiv.org/abs/2209.00840} {Folio: Natural language reasoning with first-order logic}.
\newblock \emph{Preprint}, arXiv:2209.00840.

\bibitem[{Harman(1965)}]{684604fc-4b8d-3190-bf8c-948758999eb7}
Gilbert~H. Harman. 1965.
\newblock \href {http://www.jstor.org/stable/2183532} {The inference to the best explanation}.
\newblock \emph{The Philosophical Review}, 74(1):88--95.

\bibitem[{He et~al.(2024)He, Zhang, Yan, Wu, and Chen}]{he2024ideaenhancingrulelearning}
Kaiyu He, Mian Zhang, Shuo Yan, Peilin Wu, and Zhiyu~Zoey Chen. 2024.
\newblock \href {https://arxiv.org/abs/2408.10455} {Idea: Enhancing the rule learning ability of large language model agent through induction, deduction, and abduction}.
\newblock \emph{Preprint}, arXiv:2408.10455.

\bibitem[{Hempel and Oppenheim(1948)}]{Hempel_Oppenheim_1948}
Carl~G. Hempel and Paul Oppenheim. 1948.
\newblock \href {https://doi.org/10.1086/286983} {Studies in the logic of explanation}.
\newblock \emph{Philosophy of Science}, 15(2):135–175.

\bibitem[{Higgins et~al.(2018)Higgins, Sonnerat, Matthey, Pal, Burgess, Bosnjak, Shanahan, Botvinick, Hassabis, and Lerchner}]{higgins2018scanlearninghierarchicalcompositional}
Irina Higgins, Nicolas Sonnerat, Loic Matthey, Arka Pal, Christopher~P Burgess, Matko Bosnjak, Murray Shanahan, Matthew Botvinick, Demis Hassabis, and Alexander Lerchner. 2018.
\newblock \href {https://arxiv.org/abs/1707.03389} {Scan: Learning hierarchical compositional visual concepts}.
\newblock \emph{Preprint}, arXiv:1707.03389.

\bibitem[{Hu et~al.(2022)Hu, Ma, Liu, Wei, and Bai}]{hu2022stratifiedruleawarenetworkabstract}
Sheng Hu, Yuqing Ma, Xianglong Liu, Yanlu Wei, and Shihao Bai. 2022.
\newblock \href {https://arxiv.org/abs/2002.06838} {Stratified rule-aware network for abstract visual reasoning}.
\newblock \emph{Preprint}, arXiv:2002.06838.

\bibitem[{Hu et~al.(2023)Hu, Storks, Lewis, and Chai}]{hu-etal-2023-context}
Xiaoyang Hu, Shane Storks, Richard Lewis, and Joyce Chai. 2023.
\newblock \href {https://doi.org/10.18653/v1/2023.acl-long.109} {In-context analogical reasoning with pre-trained language models}.
\newblock In \emph{Proceedings of the 61st Annual Meeting of the Association for Computational Linguistics (Volume 1: Long Papers)}, pages 1953--1969, Toronto, Canada. Association for Computational Linguistics.

\bibitem[{Jiayang et~al.(2023)Jiayang, Qiu, Chan, Fang, Wang, Chan, Ru, Guo, Zhang, Song, Zhang, and Zhang}]{jiayang2023storyanalogyderivingstorylevelanalogies}
Cheng Jiayang, Lin Qiu, Tsz~Ho Chan, Tianqing Fang, Weiqi Wang, Chunkit Chan, Dongyu Ru, Qipeng Guo, Hongming Zhang, Yangqiu Song, Yue Zhang, and Zheng Zhang. 2023.
\newblock \href {https://arxiv.org/abs/2310.12874} {Storyanalogy: Deriving story-level analogies from large language models to unlock analogical understanding}.
\newblock \emph{Preprint}, arXiv:2310.12874.

\bibitem[{Johnson-Laird(2010)}]{deductivereasoningjohnson}
Phil Johnson-Laird. 2010.
\newblock \href {https://doi.org/10.1002/wcs.20} {Deductive reasoning}.
\newblock \emph{Wiley Interdisciplinary Reviews: Cognitive Science}, 1:8 -- 17.

\bibitem[{Jones et~al.(2022)Jones, Kmiecik, Irwin, and Morrison}]{Jones2022}
Lara~L. Jones, Matthew~J. Kmiecik, Jessica~L. Irwin, and Robert~G. Morrison. 2022.
\newblock \href {https://doi.org/10.3758/s13423-022-02062-8} {Differential effects of semantic distance, distractor salience, and relations in verbal analogy}.
\newblock \emph{Psychonomic Bulletin \& Review}, 29:1480--1491.

\bibitem[{Kadiķis et~al.(2022)Kadiķis, Srivastav, and Klinger}]{kadiķis2022embarrassinglysimpleperformanceprediction}
Emīls Kadiķis, Vaibhav Srivastav, and Roman Klinger. 2022.
\newblock \href {https://arxiv.org/abs/2202.10408} {Embarrassingly simple performance prediction for abductive natural language inference}.
\newblock \emph{Preprint}, arXiv:2202.10408.

\bibitem[{Kahneman(2011)}]{kahneman2011thinking}
Daniel Kahneman. 2011.
\newblock \emph{Thinking, Fast and Slow}.
\newblock Farrar, Straus and Giroux, New York.

\bibitem[{Kim et~al.(2022)Kim, Phunyaphibarn, Ahn, and Kim}]{kim2022playgrounds}
Subin Kim, Prin Phunyaphibarn, Donghyun Ahn, and Sundong Kim. 2022.
\newblock \href {https://openreview.net/forum?id=F4RNpByoqP} {Playgrounds for abstraction and reasoning}.
\newblock In \emph{NeurIPS 2022 Workshop on Neuro Causal and Symbolic AI (nCSI)}.

\bibitem[{Li et~al.(2025)Li, Wang, Zheng, and Song}]{li-etal-2025-patterns}
Chunyang Li, Weiqi Wang, Tianshi Zheng, and Yangqiu Song. 2025.
\newblock \href {https://doi.org/10.18653/v1/2025.findings-acl.1006} {Patterns over principles: The fragility of inductive reasoning in {LLM}s under noisy observations}.
\newblock In \emph{Findings of the Association for Computational Linguistics: ACL 2025}, pages 19608--19626, Vienna, Austria. Association for Computational Linguistics.

\bibitem[{Lipton(2000)}]{Lipton2000-LIPITT-8}
Peter Lipton. 2000.
\newblock Inference to the best explanation.
\newblock In W.~Newton{-}Smith, editor, \emph{A companion to the philosophy of science}, pages 184--193. Blackwell.

\bibitem[{Liu et~al.(2024)Liu, Neubig, and Andreas}]{liu2024incompleteloopinstructioninference}
Emmy Liu, Graham Neubig, and Jacob Andreas. 2024.
\newblock \href {https://arxiv.org/abs/2404.03028} {An incomplete loop: Instruction inference, instruction following, and in-context learning in language models}.
\newblock \emph{Preprint}, arXiv:2404.03028.

\bibitem[{Mikolov et~al.(2013{\natexlab{a}})Mikolov, Chen, Corrado, and Dean}]{mikolov2013efficientestimationwordrepresentations}
Tomas Mikolov, Kai Chen, Greg Corrado, and Jeffrey Dean. 2013{\natexlab{a}}.
\newblock \href {https://arxiv.org/abs/1301.3781} {Efficient estimation of word representations in vector space}.
\newblock \emph{Preprint}, arXiv:1301.3781.

\bibitem[{Mikolov et~al.(2013{\natexlab{b}})Mikolov, Sutskever, Chen, Corrado, and Dean}]{mikolov2013distributedrepresentationswordsphrases}
Tomas Mikolov, Ilya Sutskever, Kai Chen, Greg Corrado, and Jeffrey Dean. 2013{\natexlab{b}}.
\newblock \href {https://arxiv.org/abs/1310.4546} {Distributed representations of words and phrases and their compositionality}.
\newblock \emph{Preprint}, arXiv:1310.4546.

\bibitem[{OpenAI(2024)}]{openai2024gpt4o}
OpenAI. 2024.
\newblock \href {https://openai.com/index/hello-gpt-4o/} {Hello {GPT-4o}}.

\bibitem[{{OpenAI}(2024)}]{openai2024o1preview}
{OpenAI}. 2024.
\newblock \href {https://openai.com/index/introducing-openai-o1-preview/} {Introducing openai o1 preview}.
\newblock Accessed: 2025-02-14.

\bibitem[{OpenAI(2025)}]{openai2025openaiO3Mini}
OpenAI. 2025.
\newblock \href {https://openai.com/index/openai-o3-mini/} {Openai o3 mini: Pushing the frontier of cost-effective reasoning}.

\bibitem[{Peirce(1958)}]{peirce_1958}
Charles~Sanders Peirce. 1958.
\newblock \emph{Collected Papers of Charles Sanders Peirce}, volume 1-6.
\newblock Harvard University Press, Cambridge, MA.

\bibitem[{Phillips et~al.(2016)Phillips, Fletcher, Marks, and Hine}]{phillips2016thinking}
Wendy~J. Phillips, Janet~M. Fletcher, Anthony D.~G. Marks, and Donald~W. Hine. 2016.
\newblock \href {https://doi.org/10.1037/bul0000027} {Thinking styles and decision making: A meta-analysis}.
\newblock \emph{Psychological Bulletin}, 142(3):260--290.

\bibitem[{Qin et~al.(2021)Qin, Shwartz, West, Bhagavatula, Hwang, Bras, Bosselut, and Choi}]{qin2021futureunsupervisedbackpropbaseddecoding}
Lianhui Qin, Vered Shwartz, Peter West, Chandra Bhagavatula, Jena Hwang, Ronan~Le Bras, Antoine Bosselut, and Yejin Choi. 2021.
\newblock \href {https://arxiv.org/abs/2010.05906} {Back to the future: Unsupervised backprop-based decoding for counterfactual and abductive commonsense reasoning}.
\newblock \emph{Preprint}, arXiv:2010.05906.

\bibitem[{Qiu et~al.(2024)Qiu, Jiang, Lu, Sclar, Pyatkin, Bhagavatula, Wang, Kim, Choi, Dziri, and Ren}]{qiu2024phenomenal}
Linlu Qiu, Liwei Jiang, Ximing Lu, Melanie Sclar, Valentina Pyatkin, Chandra Bhagavatula, Bailin Wang, Yoon Kim, Yejin Choi, Nouha Dziri, and Xiang Ren. 2024.
\newblock \href {https://openreview.net/forum?id=bNt7oajl2a} {Phenomenal yet puzzling: Testing inductive reasoning capabilities of language models with hypothesis refinement}.
\newblock In \emph{The Twelfth International Conference on Learning Representations}.

\bibitem[{Qwen et~al.(2025)Qwen, :, Yang, Yang, Zhang, Hui, Zheng, Yu, Li, Liu, Huang, Wei, Lin, Yang, Tu, Zhang, Yang, Yang, Zhou, Lin, Dang, Lu, Bao, Yang, Yu, Li, Xue, Zhang, Zhu, Men, Lin, Li, Tang, Xia, Ren, Ren, Fan, Su, Zhang, Wan, Liu, Cui, Zhang, and Qiu}]{qwen2025qwen25technicalreport}
Qwen, :, An~Yang, Baosong Yang, Beichen Zhang, Binyuan Hui, Bo~Zheng, Bowen Yu, Chengyuan Li, Dayiheng Liu, Fei Huang, Haoran Wei, Huan Lin, Jian Yang, Jianhong Tu, Jianwei Zhang, Jianxin Yang, Jiaxi Yang, Jingren Zhou, Junyang Lin, Kai Dang, Keming Lu, Keqin Bao, Kexin Yang, Le~Yu, Mei Li, Mingfeng Xue, Pei Zhang, Qin Zhu, Rui Men, Runji Lin, Tianhao Li, Tianyi Tang, Tingyu Xia, Xingzhang Ren, Xuancheng Ren, Yang Fan, Yang Su, Yichang Zhang, Yu~Wan, Yuqiong Liu, Zeyu Cui, Zhenru Zhang, and Zihan Qiu. 2025.
\newblock \href {https://arxiv.org/abs/2412.15115} {Qwen2.5 technical report}.
\newblock \emph{Preprint}, arXiv:2412.15115.

\bibitem[{Rafailov et~al.(2024)Rafailov, Sharma, Mitchell, Ermon, Manning, and Finn}]{rafailov2024directpreferenceoptimizationlanguage}
Rafael Rafailov, Archit Sharma, Eric Mitchell, Stefano Ermon, Christopher~D. Manning, and Chelsea Finn. 2024.
\newblock \href {https://arxiv.org/abs/2305.18290} {Direct preference optimization: Your language model is secretly a reward model}.
\newblock \emph{Preprint}, arXiv:2305.18290.

\bibitem[{Raven(1938)}]{raven1938}
J.~C. Raven. 1938.
\newblock \emph{Raven's Progressive Matrices}.
\newblock Western Psychological Services.

\bibitem[{Reiter(1987)}]{REITER198757}
Raymond Reiter. 1987.
\newblock \href {https://doi.org/10.1016/0004-3702(87)90062-2} {A theory of diagnosis from first principles}.
\newblock \emph{Artificial Intelligence}, 32(1):57--95.

\bibitem[{Ribeiro et~al.(2018)Ribeiro, Singh, and Guestrin}]{Ribeiro2018SemanticallyEA}
Marco~Tulio Ribeiro, Sameer Singh, and Carlos Guestrin. 2018.
\newblock \href {https://api.semanticscholar.org/CorpusID:21740766} {Semantically equivalent adversarial rules for debugging nlp models}.
\newblock In \emph{Annual Meeting of the Association for Computational Linguistics}.

\bibitem[{Rule(2020)}]{rule2020child}
Joshua~S Rule. 2020.
\newblock \emph{The child as hacker: {{Building}} more human-like models of learning}.
\newblock Ph.D. thesis, MIT.

\bibitem[{Salmon(1984)}]{salmon_1984}
Merrilee~H. Salmon. 1984.
\newblock \emph{Introduction to Logic and Critical Thinking}.
\newblock Harcourt Brace Jovanovich, Fort Worth.

\bibitem[{Saparov et~al.(2023)Saparov, Pang, Padmakumar, Joshi, Kazemi, Kim, and He}]{saparov2023testinggeneraldeductivereasoning}
Abulhair Saparov, Richard~Yuanzhe Pang, Vishakh Padmakumar, Nitish Joshi, Seyed~Mehran Kazemi, Najoung Kim, and He~He. 2023.
\newblock \href {https://arxiv.org/abs/2305.15269} {Testing the general deductive reasoning capacity of large language models using ood examples}.
\newblock \emph{Preprint}, arXiv:2305.15269.

\bibitem[{Simonyan and Zisserman(2015)}]{simonyan2015deepconvolutionalnetworkslargescale}
Karen Simonyan and Andrew Zisserman. 2015.
\newblock \href {https://arxiv.org/abs/1409.1556} {Very deep convolutional networks for large-scale image recognition}.
\newblock \emph{Preprint}, arXiv:1409.1556.

\bibitem[{Sourati et~al.(2024)Sourati, Ilievski, Sommerauer, and Jiang}]{sourati2024arnanalogicalreasoningnarratives}
Zhivar Sourati, Filip Ilievski, Pia Sommerauer, and Yifan Jiang. 2024.
\newblock \href {https://arxiv.org/abs/2310.00996} {Arn: Analogical reasoning on narratives}.
\newblock \emph{Preprint}, arXiv:2310.00996.

\bibitem[{Stevenson et~al.(2024)Stevenson, Pafford, Maas, and Mitchell}]{claire2024analogy}
Claire Stevenson, Alexandra Pafford, Han Maas, and Melanie Mitchell. 2024.
\newblock \href {https://doi.org/10.48550/arXiv.2411.02348} {Can large language models generalize analogy solving like people can?}

\bibitem[{Tafjord et~al.(2021)Tafjord, Mishra, and Clark}]{tafjord2021proofwritergeneratingimplicationsproofs}
Oyvind Tafjord, Bhavana~Dalvi Mishra, and Peter Clark. 2021.
\newblock \href {https://arxiv.org/abs/2012.13048} {Proofwriter: Generating implications, proofs, and abductive statements over natural language}.
\newblock \emph{Preprint}, arXiv:2012.13048.

\bibitem[{Tanzer et~al.(2024)Tanzer, Suzgun, Visser, Jurafsky, and Melas-Kyriazi}]{tanzer2024benchmarklearningtranslatenew}
Garrett Tanzer, Mirac Suzgun, Eline Visser, Dan Jurafsky, and Luke Melas-Kyriazi. 2024.
\newblock \href {https://arxiv.org/abs/2309.16575} {A benchmark for learning to translate a new language from one grammar book}.
\newblock \emph{Preprint}, arXiv:2309.16575.

\bibitem[{Thagard(2024)}]{thagard2024chatgptmakeexplanatoryinferences}
Paul Thagard. 2024.
\newblock \href {https://arxiv.org/abs/2404.18982} {Can chatgpt make explanatory inferences? benchmarks for abductive reasoning}.
\newblock \emph{Preprint}, arXiv:2404.18982.

\bibitem[{Turney(2008)}]{Turney_2008}
P.~D. Turney. 2008.
\newblock \href {https://doi.org/10.1613/jair.2693} {The latent relation mapping engine: Algorithm and experiments}.
\newblock \emph{Journal of Artificial Intelligence Research}, 33:615–655.

\bibitem[{Vendetti et~al.(2012)Vendetti, Knowlton, and Holyoak}]{Vendetti2012}
Marianna~S. Vendetti, Barbara~J. Knowlton, and Keith~J. Holyoak. 2012.
\newblock \href {https://doi.org/10.3758/s13415-012-0103-0} {The impact of semantic distance and induced stress on analogical reasoning: A neurocomputational account}.
\newblock \emph{Cognitive, Affective, \& Behavioral Neuroscience}, 12(4):804--812.

\bibitem[{Wang et~al.(2024)Wang, Zelikman, Poesia, Pu, Haber, and Goodman}]{wang2024hypothesis}
Ruocheng Wang, Eric Zelikman, Gabriel Poesia, Yewen Pu, Nick Haber, and Noah Goodman. 2024.
\newblock \href {https://openreview.net/forum?id=G7UtIGQmjm} {Hypothesis search: Inductive reasoning with language models}.
\newblock In \emph{The Twelfth International Conference on Learning Representations}.

\bibitem[{Webb et~al.(2023)Webb, Holyoak, and Lu}]{webb2023emergentanalogicalreasoninglarge}
Taylor Webb, Keith~J. Holyoak, and Hongjing Lu. 2023.
\newblock \href {https://arxiv.org/abs/2212.09196} {Emergent analogical reasoning in large language models}.
\newblock \emph{Preprint}, arXiv:2212.09196.

\bibitem[{Wei et~al.(2023)Wei, Wang, Schuurmans, Bosma, Ichter, Xia, Chi, Le, and Zhou}]{wei2023chainofthoughtpromptingelicitsreasoning}
Jason Wei, Xuezhi Wang, Dale Schuurmans, Maarten Bosma, Brian Ichter, Fei Xia, Ed~Chi, Quoc Le, and Denny Zhou. 2023.
\newblock \href {https://arxiv.org/abs/2201.11903} {Chain-of-thought prompting elicits reasoning in large language models}.
\newblock \emph{Preprint}, arXiv:2201.11903.

\bibitem[{Wijesiriwardene et~al.(2023)Wijesiriwardene, Wickramarachchi, Gajera, Gowaikar, Gupta, Chadha, Reganti, Sheth, and Das}]{wijesiriwardene2023analogicalnovelbenchmark}
Thilini Wijesiriwardene, Ruwan Wickramarachchi, Bimal~G. Gajera, Shreeyash~Mukul Gowaikar, Chandan Gupta, Aman Chadha, Aishwarya~Naresh Reganti, Amit Sheth, and Amitava Das. 2023.
\newblock \href {https://arxiv.org/abs/2305.05050} {Analogical -- a novel benchmark for long text analogy evaluation in large language models}.
\newblock \emph{Preprint}, arXiv:2305.05050.

\bibitem[{Xu et~al.(2023)Xu, Li, Vaezipoor, Sanner, and Khalil}]{xu2023llms}
Yudong Xu, Wenhao Li, Pashootan Vaezipoor, Scott Sanner, and Elias~B Khalil. 2023.
\newblock Llms and the abstraction and reasoning corpus: Successes, failures, and the importance of object-based representations.
\newblock \emph{arXiv preprint arXiv:2305.18354}.

\bibitem[{Ye et~al.(2024)Ye, Wang, Choi, Lu, Sharma, Shen, Tiyyala, Andrews, and Khashabi}]{ye2024analobenchbenchmarkingidentificationabstract}
Xiao Ye, Andrew Wang, Jacob Choi, Yining Lu, Shreya Sharma, Lingfeng Shen, Vijay Tiyyala, Nicholas Andrews, and Daniel Khashabi. 2024.
\newblock \href {https://arxiv.org/abs/2402.12370} {Analobench: Benchmarking the identification of abstract and long-context analogies}.
\newblock \emph{Preprint}, arXiv:2402.12370.

\bibitem[{Zhang et~al.(2019)Zhang, Gao, Jia, Zhu, and Zhu}]{zhang2019raven}
Chi Zhang, Feng Gao, Baoxiong Jia, Yixin Zhu, and Song-Chun Zhu. 2019.
\newblock Raven: A dataset for relational and analogical visual reasoning.
\newblock In \emph{Proceedings of the IEEE Conference on Computer Vision and Pattern Recognition (CVPR)}.

\bibitem[{Zhao et~al.(2024)Zhao, Chiu, Hwang, Brahman, Hessel, Choudhury, Choi, Li, and Suhr}]{zhao2024uncommonsensereasoningabductivereasoning}
Wenting Zhao, Justin~T Chiu, Jena~D. Hwang, Faeze Brahman, Jack Hessel, Sanjiban Choudhury, Yejin Choi, Xiang~Lorraine Li, and Alane Suhr. 2024.
\newblock \href {https://arxiv.org/abs/2311.08469} {Uncommonsense reasoning: Abductive reasoning about uncommon situations}.
\newblock \emph{Preprint}, arXiv:2311.08469.

\bibitem[{Zheng et~al.(2025{\natexlab{a}})Zheng, Chen, Li, Li, Zong, Shi, Xu, Song, Wong, and See}]{zheng2025cursecotlimitationschainofthought}
Tianshi Zheng, Yixiang Chen, Chengxi Li, Chunyang Li, Qing Zong, Haochen Shi, Baixuan Xu, Yangqiu Song, Ginny~Y. Wong, and Simon See. 2025{\natexlab{a}}.
\newblock \href {https://arxiv.org/abs/2504.05081} {The curse of cot: On the limitations of chain-of-thought in in-context learning}.
\newblock \emph{Preprint}, arXiv:2504.05081.

\bibitem[{Zheng et~al.(2025{\natexlab{b}})Zheng, Wang, Wang, Bai, Yin, Deng, Song, and Li}]{zheng-etal-2025-enhancing-transformers}
Tianshi Zheng, Jiazheng Wang, Zihao Wang, Jiaxin Bai, Hang Yin, Zheye Deng, Yangqiu Song, and Jianxin Li. 2025{\natexlab{b}}.
\newblock \href {https://doi.org/10.18653/v1/2025.acl-long.274} {Enhancing transformers for generalizable first-order logical entailment}.
\newblock In \emph{Proceedings of the 63rd Annual Meeting of the Association for Computational Linguistics (Volume 1: Long Papers)}, pages 5505--5524, Vienna, Austria. Association for Computational Linguistics.

\bibitem[{Zhong et~al.(2023)Zhong, Wei, Yang, Wu, Liu, Wei, Li, Yao, Ma, Li, Zhu, Jiang, Han, Shen, Liu, and Zhang}]{zhong2023chatablabductivelearningnatural}
Tianyang Zhong, Yaonai Wei, Li~Yang, Zihao Wu, Zhengliang Liu, Xiaozheng Wei, Wenjun Li, Junjie Yao, Chong Ma, Xiang Li, Dajiang Zhu, Xi~Jiang, Junwei Han, Dinggang Shen, Tianming Liu, and Tuo Zhang. 2023.
\newblock \href {https://arxiv.org/abs/2304.11107} {Chatabl: Abductive learning via natural language interaction with chatgpt}.
\newblock \emph{Preprint}, arXiv:2304.11107.

\bibitem[{Zong et~al.(2024)Zong, Wang, Zheng, Ren, and Song}]{zong2024comparisonqaevaluatingfactualityrobustness}
Qing Zong, Zhaowei Wang, Tianshi Zheng, Xiyu Ren, and Yangqiu Song. 2024.
\newblock \href {https://arxiv.org/abs/2412.20251} {Comparisonqa: Evaluating factuality robustness of llms through knowledge frequency control and uncertainty}.
\newblock \emph{Preprint}, arXiv:2412.20251.

\end{thebibliography}

\newpage

\appendix

\,

\newpage
\section{Model Details}
\label{app:llm}
In our experiments, we tested 15 modern LLM / MLLMs / LRMs with details as follows:
\begin{itemize}
\item \textbf{Qwen-2.5-7b / Qwen-2.5-72b} \cite{qwen2025qwen25technicalreport} is an open-source MoE LLM series, trained with 18 trillion tokens of pre-training corpus and 1 million fine-tuning examples.
\item \textbf{Llama-3.1-70b / Llama-3.1-405b} \cite{meta2024llama31} is an open-source dense LLM series, trained with 15 trillion tokens of pre-training corpus, and adopted DPO \cite{rafailov2024directpreferenceoptimizationlanguage} during its alignment stage.
\item \textbf{GPT-4o-mini / GPT-4o} \cite{openai2024gpt4o} is the latest proprietary LLM series by OpenAI prior to their reasoning models.
\item \textbf{Gemini-1.5-flash / Gemini-1.5-pro} \cite{google2024gemini} is a proprietary MoE LLM series featuring a long context window of 1 million tokens.
\item \textbf{Gemini-2.0-flash} \cite{google2024gemini2} is the latest Gemini series LLM, offering enhanced multimodal and reasoning performance.
\item \textbf{Pixtral-12b} \cite{agrawal2024pixtral12b} is a lightweight open-source multimodal LLM.
\item \textbf{Deepseek-V3} \cite{deepseekai2024deepseekv3technicalreport} is the state-of-the-art open-source LLM.
\item \textbf{Deepseek-R1} \cite{deepseekai2025deepseekr1incentivizingreasoningcapability} is the leading open-source LRM trained with reinforcement learning using a rule-based reward system.
\item \textbf{o1-mini / o1} \cite{openai2024o1preview} represents the state-of-the-art proprietary LRM series developed by OpenAI.
\item \textbf{o3-mini} \cite{openai2025openaiO3Mini} is the latest LRM by OpenAI, featured its cost-effectiveness.

\end{itemize}
The temperature for all LLMs is set to zero in our main experiments, while it is set to 0.4 during the hypothesis sampling in our scaling experiments.
\newpage
\section{Difficulty Annotation}
\label{app:difficulty}
The detailed difficulty annotation standards are presented in Table \ref{tab:difficulty}. For \textbf{EKAR} and \textbf{VASR}, we set thresholds for semantic distances to categorize the difficulty into easy, medium, and hard, ensuring comparable sizes across categories.\[
\mathrm{sem\_dist} = \frac{\mathrm{cos\_dist}(A, B) + \mathrm{cos\_dist}(A', B')}{2}
\]For \textbf{RAVEN}, we calculate the number of attributes in transition among rows, with fine-grained categorization applied within each question typology. For \textbf{List Function}, we use the predefined complexity ranking of mapping functions provided by \cite{rule2020child}. For \textbf{SALT}, we classify the syntax complexity of the translation examples into simple, medium, and complex categories.
\setlength{\tabcolsep}{8pt}

\begin{table*}[t]
\centering
\scriptsize
\begin{tabular}{cclccc}
\toprule
\textbf{Dataset} & \textbf{Determinator} & \multicolumn{1}{c}{\textbf{Category}} & \textbf{Easy} & \textbf{Medium} & \textbf{Hard} \\ \midrule
E-KAR & FastText Distance & \multicolumn{1}{c}{-} & \textless{}0.70 & 0.70$\sim$0.80 & \textgreater{}0.80 \\ \midrule
VASR & VGG Distance & \multicolumn{1}{c}{-} & \textless{}0.70 & 0.70$\sim$0.76 & \textgreater{}0.76 \\ \midrule
\multirow{7}{*}{RAVEN} & \multirow{7}{*}{Number of Transitions} & center\_single & 1 & 2 & \textgreater{}=3 \\
 &  & distribute\_four & \textless{}=2 & 3 & \textgreater{}=4 \\
 &  & distribute\_nine & \textless{}=2 & 3 & \textgreater{}=4 \\
 &  & in\_center\_single\_out\_center\_single & \textless{}=3 & 4 & \textgreater{}=5 \\
 &  & in\_distribute\_four\_out\_center\_single & \textless{}=3 & 4 & \textgreater{}=5 \\
 &  & up\_center\_single\_down\_center\_single & \textless{}=3 & 4 & \textgreater{}=5 \\
 &  & left\_center\_single\_right\_center\_single & \textless{}=4 & 5 & \textgreater{}=6 \\ \midrule
List Function & Function Complexity Ranking & \multicolumn{1}{c}{-} & \textless{}=84 & 85$\sim$170 & \textgreater{}=170 \\ \midrule
SALT & Syntax Complexity & \multicolumn{1}{c}{-} & simple & intermediate & complex \\ \bottomrule
\end{tabular}
\caption{Difficulty classification standards for each datasets in our experiment.}
\label{tab:difficulty}

\end{table*}
\begin{table*}[]
\centering
\scriptsize
\begin{tabular}{ccc}
\toprule
\textbf{English Sentence} & \textit{I like beautiful house.} & \textit{Giant elephant runs quickly.} \\ \midrule
\textbf{Syntax Structure} & \textless{}pronoun - verb - adjective - noun\textgreater{} & \textless{}adjective - noun - verb - adverb\textgreater{} \\ \midrule
\textbf{Grammar Rule} & \textless{}noun-adjective inversion\textgreater{} & \textless{}predicate-subject inversion\textgreater{} \\ \midrule
\textbf{Transition Type} & Intra-Constituent & Inter-Constituent \\ \midrule
\textbf{Vocabulary} & I \(\rightarrow\) gkt, like \(\rightarrow\) ivo, beautiful \(\rightarrow\) prr, house \(\rightarrow\) cbi & giant \(\rightarrow\) rgd, elephant \(\rightarrow\) krt, runs \(\rightarrow\) uco, quickly \(\rightarrow\) xrk \\ \midrule
\textbf{Translation} & \textit{gkt ivo cbi prr.} & \textit{uco xrk rgd krt.} \\ \bottomrule
\end{tabular}
\caption{Examples of intra-constituent and inter-constituent syntactic transitions in the SALT dataset. }
\label{tab:salt_example}
\end{table*}
\newpage
\section{Syntax-aware Artificial Language Translation}
\vspace{-0.1cm}
\label{app:salt}
Syntax-aware Artificial Language Translation (SALT) is a low-resource machine translation (MT) benchmark that we designed and developed to evaluate generalizable in-context learning in large language models. LLMs are required to infer vocabulary mappings as well as syntactic transitions from few-shot demonstrations and apply them to translate a compositionally crafted testing instance. SALT offers two key advantages over other low-resource MT benchmarks: 1) SALT synthesizes out-of-vocabulary strings for the artificial language, \textbf{preventing data leakage}, a common issue in other benchmarks.  2) SALT provides \textbf{detailed difficulty control} enabled by human-curated syntactic structures with compositional complexities.  

The creation of SALT involves two main stages:  
\begin{enumerate}
    \item \textbf{Syntax-aware Template Design} In the first stage, we design syntactic rules that involve the permutation or repetition of semantic units in the artificial language, as illustrated in Table \ref{tab:salt_example} and \ref{tab:salt-modes}. Next, we manually craft templates for few-shot demonstrations with considerations in compositional generalization. We ensure that all the necessary underlying word mappings and syntactic rules required for translating the testing instances can be inferred from the provided few-shot demonstrations.
    \item \textbf{Semantic-aware Data Synthesis} After acquiring the templates, we populate them with semantically appropriate English words using LLM-assisted selection. Next, we randomly assign out-of-vocabulary letter strings as the artificial language equivalents for each English word. Finally, a total of 1,200 questions are sampled—400 at each difficulty level—ensuring comparability in size with other datasets.
\end{enumerate}
\begin{table}[]
\small
\begin{tabular}{p{0.13\linewidth} p{0.44\linewidth} p{0.22\linewidth}}
\toprule
\textbf{Difficulty Level} & \textbf{Syntax Rule} & \textbf{Sentence Complexity} \\
\midrule
Easy  & word-to-word mapping & Simple \\
Easy  & noun repetition & Simple \\
Easy  & noun-adjective inversion & Simple \\
Easy  & predicate-subject inversion & Simple \\
Medium  & word-to-word mapping & Intermediate \\
Medium  & verb repetition & Intermediate \\
Medium  & noun-adjective inversion & Intermediate \\
Medium  & predicate-subject inversion & Intermediate \\
Hard  & word-to-word mapping & Complex \\
Hard  & adjective repetition & Complex \\
Hard  & verb-adverb inversion & Complex \\
Hard  & predicate-subject inversion & Complex \\
\bottomrule
\end{tabular}
\caption{Summary of the 12 translation modes in SALT, listing their syntax rule and sentence complexity.}
\label{tab:salt-modes}
\end{table}


\newpage






\newpage
\section{Prompt Templates}
\label{app:prompt}
\begin{promptbox}[colback=pink!10, colframe=pink!50!purple, title=Textual Analogy (Induction)]{}
\scriptsize
Below is an analogy question, where analogy x:y::x':y' exists between the two wordsets, your task is to finish the second wordset to complete the analogy. \\
\begin{verbatim}
Wordset1: <word_x>:<word_x'>
Wordset2: <word_y>:[missing_word]
  
Your response should strictly follow the JSON dict format:

{
    "answer": "missing word here"
}
\end{verbatim}
\end{promptbox}

\begin{promptbox}[colback=pink!10, colframe=pink!50!purple, title=Textual Analogy (Automatic)]{}
\scriptsize
Below is an analogy question, where analogy x:y::x':y' exists between the two wordsets, your task is to finish the second wordset to complete the analogy. \\
\begin{verbatim}
Wordset1: <word_x>:<word_x'>
Wordset2: <word_y>:[missing_word]
  
Your response should strictly follow the JSON dict format:

{
    "reasoning":"reasoning steps here",
    "answer": "missing word here"
}
\end{verbatim}
\end{promptbox}
\newpage
\begin{promptbox}[colback=pink!10, colframe=pink!50!purple, title=Textual Analogy (Abduction)]{}
\scriptsize
Below is an analogy question, where analogy x:y::x':y' exists between the two wordsets, your task is to infer the relational pattern within wordsets. \\
\begin{verbatim}
Wordset1: <word_x>:<word_x'>
Wordset2: <word_y>:[missing_word]
  
Your response should strictly follow the JSON dict format:

{
    "reasoning": "reasoning steps here"
    "pattern": "relational pattern here"
}
\end{verbatim}
\end{promptbox}

\begin{promptbox}[colback=pink!10, colframe=pink!50!purple, title=Textual Analogy (Deduction)]{}
\scriptsize
Below is an analogy question, where analogy x:y::x':y' exists between the two wordsets, your task is to finish the second wordset to complete the analogy. Here's the relational pattern: <pattern> \\
\begin{verbatim}
Wordset1: <word_x>:<word_x'>
Wordset2: <word_y>:[missing_word]
  
Your response should strictly follow the JSON dict format:

{
    "reasoning":"reasoning steps here",
    "answer": "missing word here"
}
\end{verbatim}
\end{promptbox}

\newpage
\begin{promptbox}[colback=green!10, colframe=green!60!black, title=Visual Analogy (Induction)]{}
\scriptsize
Below is an analogy question, where analogy x:y::x':y' exists between the two image pairs, your task is to complete the second image pair to complete the analogy. \\
\begin{verbatim}
Image Pair 1: <img_x>:<img_x'>
Image Pair 2: <img_y>:[missing_img]

<Candidate Images>

Your response should strictly follow the JSON dict format:

{
    "answer": "missing image choice here"
}
\end{verbatim}
\end{promptbox}

\begin{promptbox}[colback=green!10, colframe=green!60!black, title=Visual Analogy (Automatic)]{}
\scriptsize
Below is an analogy question, where analogy x:y::x':y' exists between the two image pairs, your task is to complete the second image pair to complete the analogy. \\
\begin{verbatim}
Image Pair 1: <img_x>:<img_x'>
Image Pair 2: <img_y>:[missing_img]

<Candidate Images>

Your response should strictly follow the JSON dict format:

{
    "reasoning":"reasoning steps here",
    "answer": "missing image choice here"
}
\end{verbatim}
\end{promptbox}

\begin{promptbox}[colback=green!10, colframe=green!60!black, title=Visual Analogy (Abduction)]{}
\scriptsize
Below is an analogy question, where analogy x:y::x':y' exists between the two image pairs, your task is to infer the relational pattern within image pairs. \\
\begin{verbatim}
Image Pair 1: <img_x>:<img_x'>
Image Pair 2: <img_y>:[missing_img]

<Candidate Images>

Your response should strictly follow the JSON dict format:

{
    "reasoning":"reasoning steps here",
    "pattern": "relational pattern here"
}
\end{verbatim}
\end{promptbox}

\begin{promptbox}[colback=green!10, colframe=green!60!black, title=Visual Analogy (Deduction)]{}
\scriptsize
Below is an analogy question, where analogy x:y::x':y' exists between the two image pairs, your task is to complete the second image pair to complete the analogy. Here's the relational pattern: <pattern> \\
\begin{verbatim}
Image Pair 1: <img_x>:<img_x'>
Image Pair 2: <img_y>:[missing_img]

<Candidate Images>

Your response should strictly follow the JSON dict format:

{
    "reasoning":"reasoning steps here",
    "answer": "missing image choice here"
}
\end{verbatim}
\end{promptbox}
\newpage
\begin{promptbox}[colback=blue!10, colframe=blue!60!black, title=Symbolic Analogy (Induction)]{}
\scriptsize
Below is a 3x3 matrix of abstracted symbols. The symbols follow a certain rule or pattern in rows. Your task is to infer the missing symbol. \\
\begin{verbatim}
Incomplete Matrix: <incomplete_matrix>

Your response should strictly follow the JSON dict format:

{
    "answer": "missing symbol here"
}
\end{verbatim}
\end{promptbox}

\begin{promptbox}[colback=blue!10, colframe=blue!60!black, title=Symbolic Analogy (Automatic)]{}
\scriptsize
Below is a 3x3 matrix of abstracted symbols. The symbols follow a certain rule or pattern in rows. Your task is to infer the missing symbol. \\
\begin{verbatim}
Incomplete Matrix: <incomplete_matrix>

Your response should strictly follow the JSON dict format:

{
    "reasoning":"reasoning steps here",
    "answer": "missing symbol here"
}
\end{verbatim}
\end{promptbox}

\begin{promptbox}[colback=blue!10, colframe=blue!60!black, title=Symbolic Analogy (Abduction)]{}
\scriptsize
Below is a 3x3 matrix of abstracted symbols. The symbols follow a certain rule or pattern in rows. Your task is to infer the relational pattern. \\
\begin{verbatim}
Incomplete Matrix: <incomplete_matrix>

Your response should strictly follow the JSON dict format:

{
    "reasoning":"reasoning steps here",
    "pattern": "relational pattern here"
}
\end{verbatim}
\end{promptbox}

\begin{promptbox}[colback=blue!10, colframe=blue!60!black, title=Symbolic Analogy (Deduction)]{}
\scriptsize
Below is a 3x3 matrix of abstracted symbols. The symbols follow a certain rule or pattern in rows. Your task is to infer the missing symbol. Here's the relational pattern: <pattern>\\
\begin{verbatim}
Incomplete Matrix: <incomplete_matrix>

Your response should strictly follow the JSON dict format:

{
    "reasoning":"reasoning steps here",
    "answer": "missing symbol here"
}
\end{verbatim}
\end{promptbox}

\newpage

\begin{promptbox}[colback=yellow!10, colframe=yellow!40!brown, title=List Function ICL (Induction)]{}
\scriptsize
Below are several examples of input and output lists. There exists an unified function that maps the input list to the output list. \\
\begin{verbatim}

Input 1: <input_list1>, Output 1: <output_list1>
Input 2: <input_list2>, Output 2: <output_list2>
Input 3: <input_list3>, Output 3: <output_list3>

Please infer the output list for the new input list below:
New Input: <new_input_list>

Your response should strictly follow the JSON dict format:

{
    "answer": "output list here"
}
\end{verbatim}
\end{promptbox}

\begin{promptbox}[colback=yellow!10, colframe=yellow!40!brown, title=List Function ICL (Automatic)]{}
\scriptsize
Below are several examples of input and output lists. There exists an unified function that maps the input list to the output list. \\
\begin{verbatim}

Input 1: <input_list1>, Output 1: <output_list1>
Input 2: <input_list2>, Output 2: <output_list2>
Input 3: <input_list3>, Output 3: <output_list3>

Please infer the output list for the new input list below:
New Input: <new_input_list>

Your response should strictly follow the JSON dict format:

{
    "reasoning":"reasoning steps here",
    "answer": "output list here"
}
\end{verbatim}
\end{promptbox}

\begin{promptbox}[colback=yellow!10, colframe=yellow!40!brown, title=List Function ICL (Abduction)]{}
\scriptsize
Below are several examples of input and output lists. There exists an unified function that maps the input list to the output list. \\
\begin{verbatim}

Input 1: <input_list1>, Output 1: <output_list1>
Input 2: <input_list2>, Output 2: <output_list2>
Input 3: <input_list3>, Output 3: <output_list3>

Please infer the mapping function in python.
Your response should strictly follow the JSON dict format:

{
    "reasoning":"reasoning steps here",
    "function": "python function here"
}
\end{verbatim}
\end{promptbox}

\begin{promptbox}[colback=yellow!10, colframe=yellow!40!brown, title=List Function ICL (Deduction)]{}
\scriptsize
Below are several examples of input and output lists. There exists an unified function that maps the input list to the output list. The python code for the function is: <function> \\
\begin{verbatim}

Input 1: <input_list1>, Output 1: <output_list1>
Input 2: <input_list2>, Output 2: <output_list2>
Input 3: <input_list3>, Output 3: <output_list3>

Please infer the output list for the new input list below:
New Input: <new_input_list>

Your response should strictly follow the JSON dict format:

{
    "reasoning":"reasoning steps here",
    "answer": "output list here"
}
\end{verbatim}
\end{promptbox}

\newpage

\begin{promptbox}[colback=blue!20!green!5, colframe=green!55!blue!80, title=SALT ICL (Induction)]{}
\scriptsize
You are required to translate english sentences to an artificial language.
The translation involves both vocabulary mapping and syntax rules transition. Below are translation examples:\\
\begin{verbatim}

English 1: <english_1>, Translation 1: <translation_1>
English 2: <english_2>, Translation 2: <translation_2>
English 3: <english_3>, Translation 3: <translation_3>
English 4: <english_4>, Translation 4: <translation_4>

Please translate this sentence: <english_new>
Your response should strictly follow the JSON dict format:
{
    "translation": "translated sentence here"
}
\end{verbatim}
\end{promptbox}

\begin{promptbox}[colback=blue!20!green!5, colframe=green!55!blue!80, title=SALT ICL (Automatic)]{}
\scriptsize
You are required to translate english sentences to an artificial language.
The translation involves both vocabulary mapping and syntax rules transition. Below are translation examples:\\
\begin{verbatim}

English 1: <english_1>, Translation 1: <translation_1>
English 2: <english_2>, Translation 2: <translation_2>
English 3: <english_3>, Translation 3: <translation_3>
English 4: <english_4>, Translation 4: <translation_4>

Please translate this sentence: <english_new>
Your response should strictly follow the JSON dict format:
{
    "reasoning":"reasoning steps here",
    "translation": "translated sentence here"
}
\end{verbatim}
\end{promptbox}

\begin{promptbox}[colback=blue!20!green!5, colframe=green!55!blue!80, title=SALT ICL (Abduction)]{}
\scriptsize
You are required to study translations from english sentences to an artificial language.
The translation involves both vocabulary mapping and syntax rules transition. Below are translation examples:\\
\begin{verbatim}

English 1: <english_1>, Translation 1: <translation_1>
English 2: <english_2>, Translation 2: <translation_2>
English 3: <english_3>, Translation 3: <translation_3>
English 4: <english_4>, Translation 4: <translation_4>

Please infer the word mappings and syntax rules.
Your response should strictly follow the JSON dict format:
{
    "reasoning":"reasoning steps here",
    "vocabulary": "word mappings here",
    "grammar": "syntax rules here"
}
\end{verbatim}
\end{promptbox}

\begin{promptbox}[colback=blue!20!green!5, colframe=green!55!blue!80, title=SALT ICL (Deduction)]{}
\scriptsize
You are required to translate english sentences to an artificial language.
The translation involves both vocabulary mapping and syntax rules transition. Vocabulary mapping: <vocab>; Syntax rules: <grammar>. Below are translation examples:\\
\begin{verbatim}

English 1: <english_1>, Translation 1: <translation_1>
English 2: <english_2>, Translation 2: <translation_2>
English 3: <english_3>, Translation 3: <translation_3>
English 4: <english_4>, Translation 4: <translation_4>

Please translate this sentence: <english_new>
Your response should strictly follow the JSON dict format:
{
    "reasoning":"reasoning steps here",
    "translation": "translated sentence here"
}
\end{verbatim}
\end{promptbox}
\section{Interpretation on Task-Format Dependency}
\label{app:interpret}
We investigated System 2's limitations in free-text generation using the \textit{List Function} dataset, where intermediate rules are evaluatable Python functions. This allows direct assessment of abductive inference accuracy. We compared the accuracy of LLMs generating these Python functions (abduction) against their accuracy in applying ground-truth functions (deduction).

As evidenced by the results in Table \ref{tab:adcompare}, the substantially lower abduction accuracy indicates that a primary reason for System 2's failure in free-text rule execution ICL is the insufficient ability of LLMs to precisely generate rules.

\begin{table}[h!]
\centering
\small
\begin{tabular}{lcc} 
\toprule
\textbf{Model}           & \textbf{Abduction} & \textbf{Deduction} \\ \midrule 
Qwen-2.5-7b     & 26.80     & 86.64     \\
Qwen-2.5-72b    & 50.20     & 90.72     \\
GPT-4o-mini     & 40.60     & 92.56     \\ \midrule
\textbf{Average} & \textbf{39.20} & \textbf{89.97} \\ \bottomrule
\end{tabular}
\caption{Abduction vs. Deduction Accuracy (\%) on List Function Dataset.}
\label{tab:adcompare}
\end{table}

\setlength{\tabcolsep}{2.5pt}

Furthermore, to assess the impact of contextual distance from lengthy reasoning chains, characteristic of System 2 and Automatic (CoT) pipelines, we introduced dummy reasoning tokens of varying lengths before the answer in Direct Induction and Automatic pipelines on the List Function dataset (FTG). This simulates how extended context might impair free-text generation precision.

As evidenced by the results in Table \ref{tab:distance}, performance degrades for both pipelines as token length increases. This suggests that lengthy rationales contribute to task-format sensitivity by disrupting precise free-text output.

\begin{table}[h!]
\centering
\small
\begin{tabular}{cccc}
\toprule
\textbf{Pipeline} & \textbf{\begin{tabular}[c]{@{}c@{}}Contextual\\ Distance\end{tabular}} & \textbf{Qwen-2.5-7b} & \textbf{Qwen-2.5-72b} \\ \midrule
\multirow{3}{*}{\begin{tabular}[c]{@{}c@{}}Direct\\ Induction\end{tabular}} & 0 & 25.6 & 47.6 \\
 & 100 & 10.4 & 46.0 \\
 & 400 & 9.2 & 40.4 \\ \midrule
\multirow{3}{*}{\begin{tabular}[c]{@{}c@{}}Automatic\\ (Zero-shot CoT)\end{tabular}} & 0 & 27.6 & 46.8 \\
 & 100 & 17.6 & 43.6 \\
 & 400 & 16.4 & 38.8 \\ \bottomrule
\end{tabular}
\caption{Impact of Dummy Reasoning Tokens on Performance (\%) in List Function (FTG).}
\label{tab:distance}
\end{table}

\newpage
\section{Details of Scaling Experiments}
\label{app:scale}
This appendix outlines the methodologies for the scaling experiments in Section 6.
\begin{itemize}
    \item \textbf{Figure 4a (Hypothesis Selection):} The LLM first samples multiple candidate hypotheses, ranging from 1 to 10 candidates, using a temperature of 0.4. From these candidates, the LLM then selects the single best hypothesis. 

    \item \textbf{Figure 4b (Hypothesis Verification and Refinement):} Initially, a single hypothesis is obtained through the regular abduction process. This hypothesis is then subjected to iterative verification and refinement by the LLM, with this process repeated for multiple rounds. 

    \item \textbf{Figure 4c (Combined Selection and Refinement):} This approach begins with the LLM selecting the best hypothesis from several sampled candidates. The chosen hypothesis then undergoes iterative verification and refinement over multiple rounds.

    \item \textbf{Table 4 (Adaptive Scaling for DeepSeek-V3):} This method also combines selection and refinement, but with the LLM autonomously determining the number of refinement rounds within predefined limits. For the \textit{Low Consumption} setting, the LLM selects from 3 candidate hypotheses and refines the chosen one for at most 3 rounds. For the \textit{High Consumption} setting, selection is from 5 candidates, followed by refinement for at most 5 rounds.
\end{itemize}

\section{Full Results}
\label{app:full_result}
The detailed LLM performances in our analogy environement and in-context learning benchmarks is presented in tables below:
\begin{itemize}
    \item Table \ref{tab:textual-mcq}: Textual Analogy (E-KAR)-MCQ
    \item Table \ref{tab:visual-mcq}: Visual Analogy (VASR)-MCQ
    \item Table \ref{tab:symbolic-mcq}: Symbolic Analogy (RAVEN)-MCQ
    \item Table \ref{tab:textual-ftg}: Textual Analogy (E-KAR)-FTG
    \item Table \ref{tab:symbolic-ftg}: Symbolic Analogy (RAVEN)-FTG
    \item Table \ref{tab:list-mcq}: List Function ICL-MCQ
    \item Table \ref{tab:list-ftg}: List Function ICL-FTG
    \item Table \ref{tab:salt-mcq}: SALT ICL-MCQ
    \item Table \ref{tab:salt-ftg}: SALT ICL-FTG
\end{itemize}
\setlength{\tabcolsep}{12pt}

\begin{table*}[]
\centering
\scriptsize
\begin{tabular}{clcccc}
\toprule
\textbf{Model} & \multicolumn{1}{c}{\textbf{Pipeline}} & \textbf{Easy} & \textbf{Medium} & \textbf{Hard} & \textbf{Total} \\ \midrule
\multirow{3}{*}{Qwen-2.5-7b} & Induction & 65.93 & 56.32 & 40.93 & 52.64 \\
 & Automatic & 68.45 & 52.87 & 39.11 & 51.36 \\
 & Abduction+Deduction & 69.40 & 54.71 & 44.35 & 54.33 \\ \midrule
\multirow{3}{*}{Qwen-2.5-72b} & Induction & 76.03 & 68.74 & 46.77 & 61.86 \\
 & Automatic & 75.39 & 67.13 & 49.60 & 62.26 \\
 & Abduction+Deduction & 76.97 & 70.34 & 51.01 & 64.34 \\ \midrule
\multirow{3}{*}{Llama-3.1-70b} & Induction & 64.67 & 56.32 & 37.90 & 51.12 \\
 & Automatic & 73.19 & 64.14 & 46.37 & 59.38 \\
 & Abduction+Deduction & 73.19 & 62.53 & 46.17 & 58.73 \\ \midrule
\multirow{3}{*}{Llama-3.1-405b} & Induction & 74.76 & 64.83 & 43.95 & 59.05 \\
 & Automatic & 77.92 & 68.97 & 52.62 & 64.74 \\
 & Abduction+Deduction & 73.50 & 67.13 & 50.60 & 62.18 \\ \midrule
\multirow{3}{*}{GPT-4o-mini} & Induction & 66.88 & 54.94 & 36.49 & 50.64 \\
 & Automatic & 63.72 & 55.40 & 40.32 & 51.52 \\
 & Abduction+Deduction & 63.41 & 56.78 & 40.73 & 52.08 \\ \midrule
\multirow{3}{*}{GPT-4o} & Induction & 73.82 & 64.83 & 44.15 & 58.89 \\
 & Automatic & 69.72 & 63.22 & 48.59 & 59.05 \\
 & Abduction+Deduction & 73.50 & 68.74 & 51.61 & 63.14 \\ \bottomrule
\end{tabular}
\caption{LLM performances on textual analogy dataset (E-KAR) in MCQ task format.}
\label{tab:textual-mcq}
\end{table*}

\begin{table*}[]
\centering
\scriptsize
\begin{tabular}{clcccc}
\toprule
\textbf{Model} & \multicolumn{1}{c}{\textbf{Pipeline}} & \textbf{Easy} & \textbf{Medium} & \textbf{Hard} & \textbf{Total} \\ \midrule
\multirow{3}{*}{Gemini-1.5-flash} & Induction & 38.90 & 30.59 & 29.38 & 33.11 \\
 & Automatic & 54.07 & 49.83 & 47.50 & 50.71 \\
 & Abduction+Deduction & 59.34 & 47.73 & 48.75 & 51.89 \\ \midrule
\multirow{3}{*}{Gemini-1.5-pro} & Induction & 50.55 & 45.28 & 43.13 & 46.55 \\
 & Automatic & 65.49 & 54.37 & 59.06 & 59.24 \\
 & Abduction+Deduction & 65.71 & 57.34 & 59.38 & 60.65 \\ \midrule
\multirow{3}{*}{Gemini-2.0-flash} & Induction & 52.31 & 47.38 & 47.50 & 49.07 \\
 & Automatic & 63.96 & 60.84 & 61.56 & 62.06 \\
 & Abduction+Deduction & 67.47 & 59.44 & 65.62 & 63.62 \\ \midrule
\multirow{3}{*}{Pixtral-12b} & Induction & 32.53 & 24.30 & 17.81 & 25.54 \\
 & Automatic & 33.85 & 32.87 & 30.94 & 32.74 \\
 & Abduction+Deduction & 41.54 & 39.34 & 35.31 & 39.12 \\ \midrule
\multirow{3}{*}{GPT-4o-mini} & Induction & 34.73 & 26.57 & 25.31 & 29.03 \\
 & Automatic & 51.43 & 41.61 & 40.00 & 44.54 \\
 & Abduction+Deduction & 51.21 & 41.43 & 44.06 & 45.36 \\ \midrule
\multirow{3}{*}{GPT-4o} & Induction & 54.95 & 47.90 & 46.56 & 49.96 \\
 & Automatic & 66.37 & 55.07 & 59.06 & 59.84 \\
 & Abduction+Deduction & 68.13 & 59.97 & 60.94 & 62.95 \\ \bottomrule
\end{tabular}
\caption{LLM performances on visual analogy dataset (VASR) in MCQ task format.}
\label{tab:visual-mcq}
\end{table*}

\begin{table*}[]
\centering
\scriptsize
\begin{tabular}{clcccc}
\toprule
\textbf{Model} & \multicolumn{1}{c}{\textbf{Pipeline}} & \textbf{Easy} & \textbf{Medium} & \textbf{Hard} & \textbf{Total} \\ \midrule
\multirow{3}{*}{Qwen-2.5-7b} & Induction & 29.10 & 19.26 & 11.39 & 19.94 \\
 & Automatic & 29.10 & 20.35 & 14.43 & 21.29 \\
 & Abduction+Deduction & 30.10 & 21.43 & 16.46 & 22.64 \\ \midrule
\multirow{3}{*}{Qwen-2.5-72b} & Induction & 40.55 & 28.57 & 18.99 & 29.39 \\
 & Automatic & 51.24 & 38.74 & 26.08 & 38.76 \\
 & Abduction+Deduction & 54.48 & 43.72 & 36.20 & 44.80 \\ \midrule
\multirow{3}{*}{Llama-3.1-70b} & Induction & 38.06 & 28.35 & 18.73 & 28.44 \\
 & Automatic & 52.99 & 36.58 & 28.35 & 39.24 \\
 & Abduction+Deduction & 49.50 & 36.36 & 34.43 & 39.95 \\ \midrule
\multirow{3}{*}{Llama-3.1-405b} & Induction & 54.23 & 38.10 & 25.06 & 39.16 \\
 & Automatic & 53.48 & 38.53 & 28.35 & 40.11 \\
 & Abduction+Deduction & 64.93 & 47.62 & 37.72 & 50.04 \\ \midrule
\multirow{3}{*}{GPT-4o-mini} & Induction & 36.82 & 22.51 & 15.44 & 24.86 \\
 & Automatic & 37.56 & 26.41 & 12.91 & 25.73 \\
 & Abduction+Deduction & 36.32 & 25.11 & 22.78 & 27.96 \\ \midrule
\multirow{3}{*}{GPT-4o} & Induction & 41.79 & 29.87 & 17.22 & 29.71 \\
 & Automatic & 58.21 & 41.13 & 35.44 & 44.80 \\
 & Abduction+Deduction & 55.47 & 35.93 & 31.39 & 40.75 \\ \bottomrule
\end{tabular}
\caption{LLM performances on symbolic analogy dataset (RAVEN) in MCQ task format.}
\label{tab:symbolic-mcq}
\end{table*}

\begin{table*}[]
\centering
\scriptsize
\begin{tabular}{clcccc}
\toprule
\textbf{Model} & \multicolumn{1}{c}{\textbf{Pipeline}} & \textbf{Easy} & \textbf{Medium} & \textbf{Hard} & \textbf{Total} \\ \midrule
\multirow{3}{*}{Qwen-2.5-7b} & Induction & 28.08 & 22.99 & 16.33 & 21.63 \\
 & Automatic & 28.71 & 25.98 & 19.56 & 24.12 \\
 & Abduction+Deduction & 27.76 & 23.68 & 17.74 & 22.36 \\ \midrule
\multirow{3}{*}{Qwen-2.5-72b} & Induction & 35.02 & 29.20 & 21.77 & 27.72 \\
 & Automatic & 33.75 & 28.74 & 22.18 & 27.40 \\
 & Abduction+Deduction & 34.07 & 29.66 & 21.98 & 27.72 \\ \midrule
\multirow{3}{*}{Llama-3.1-70b} & Induction & 29.02 & 22.07 & 15.32 & 21.15 \\
 & Automatic & 32.81 & 23.45 & 19.15 & 24.12 \\
 & Abduction+Deduction & 29.97 & 25.52 & 18.95 & 24.04 \\ \midrule
\multirow{3}{*}{Llama-3.1-405b} & Induction & 28.71 & 24.60 & 16.94 & 22.60 \\
 & Automatic & 29.34 & 25.75 & 18.75 & 23.88 \\
 & Abduction+Deduction & 32.18 & 27.59 & 19.76 & 25.64 \\ \midrule
\multirow{3}{*}{GPT-4o-mini} & Induction & 28.08 & 22.99 & 16.33 & 21.63 \\
 & Automatic & 29.34 & 25.29 & 20.16 & 24.28 \\
 & Abduction+Deduction & 29.97 & 25.75 & 19.15 & 24.20 \\ \midrule
\multirow{3}{*}{GPT-4o} & Induction & 32.81 & 26.90 & 19.35 & 25.40 \\
 & Automatic & 32.18 & 26.21 & 20.77 & 25.56 \\
 & Abduction+Deduction & 31.23 & 27.59 & 20.36 & 25.64 \\ \bottomrule
\end{tabular}
\caption{LLM performances on textual analogy dataset (E-KAR) in FTG task format.}
\label{tab:textual-ftg}
\vspace{3cm}
\end{table*}

\begin{table*}[]
\centering
\scriptsize
\begin{tabular}{clcccc}
\toprule
\textbf{Model} & \multicolumn{1}{c}{\textbf{Pipeline}} & \textbf{Easy} & \textbf{Medium} & \textbf{Hard} & \textbf{Total} \\ \midrule
\multirow{3}{*}{Qwen-2.5-7b} & Induction & 19.15 & 8.87 & 5.57 & 11.12 \\
 & Automatic & 0.75 & 0.87 & 0.00 & 0.56 \\
 & Abduction+Deduction & 1.00 & 2.60 & 1.77 & 1.83 \\ \midrule
\multirow{3}{*}{Qwen-2.5-72b} & Induction & 37.81 & 20.13 & 13.42 & 23.67 \\
 & Automatic & 17.91 & 5.41 & 1.52 & 8.18 \\
 & Abduction+Deduction & 25.37 & 13.85 & 8.86 & 15.97 \\ \midrule
\multirow{3}{*}{Llama-3.1-70b} & Induction & 30.35 & 13.20 & 8.10 & 17.08 \\
 & Automatic & 18.16 & 7.14 & 4.81 & 9.93 \\
 & Abduction+Deduction & 9.45 & 7.58 & 6.08 & 7.70 \\ \midrule
\multirow{3}{*}{Llama-3.1-405b} & Induction & 42.29 & 20.78 & 13.42 & 25.34 \\
 & Automatic & 30.85 & 12.99 & 7.34 & 16.92 \\
 & Abduction+Deduction & 28.61 & 16.45 & 12.15 & 18.98 \\ \midrule
\multirow{3}{*}{GPT-4o-mini} & Induction & 26.37 & 12.34 & 8.61 & 15.65 \\
 & Automatic & 11.19 & 4.76 & 2.53 & 6.12 \\
 & Abduction+Deduction & 11.69 & 6.93 & 3.54 & 7.39 \\ \midrule
\multirow{3}{*}{GPT-4o} & Induction & 37.81 & 18.40 & 10.89 & 22.24 \\
 & Automatic & 16.17 & 9.09 & 5.82 & 10.33 \\
 & Abduction+Deduction & 25.12 & 14.07 & 9.37 & 16.12 \\ \bottomrule
\end{tabular}
\caption{LLM performances on symbolic analogy dataset (RAVEN) in FTG task format.}
\vspace{3cm}
\label{tab:symbolic-ftg}
\end{table*}

\begin{table*}[]
\centering
\scriptsize
\begin{tabular}{clcccc}
\toprule
\textbf{Model} & \multicolumn{1}{c}{\textbf{Pipeline}} & \textbf{Easy} & \textbf{Medium} & \textbf{Hard} & \textbf{Total} \\ \midrule
\multirow{3}{*}{Qwen-2.5-7b} & Induction & 47.69 & 33.57 & 29.87 & 37.28 \\
 & Automatic & 60.42 & 44.21 & 39.24 & 48.24 \\
 & Abduction+Deduction & 64.81 & 32.97 & 38.23 & 49.36 \\ \midrule
\multirow{3}{*}{Qwen-2.5-72b} & Induction & 65.05 & 46.34 & 40.51 & 50.96 \\
 & Automatic & 69.44 & 52.25 & 44.81 & 55.84 \\
 & Abduction+Deduction & 68.52 & 49.41 & 45.57 & 54.80 \\ \midrule
\multirow{3}{*}{GPT-4o-mini} & Induction & 59.03 & 45.86 & 33.92 & 46.64 \\
 & Automatic & 61.81 & 53.90 & 42.28 & 52.96 \\
 & Abduction+Deduction & 66.20 & 52.01 & 44.80 & 54.64 \\ \bottomrule
\end{tabular}
\caption{LLM performances on List Function dataset in MCQ task format.}
\label{tab:list-mcq}
\end{table*}

\begin{table*}[]
\centering
\scriptsize
\begin{tabular}{clcccc}
\toprule
\textbf{Model} & \multicolumn{1}{c}{\textbf{Pipeline}} & \textbf{Easy} & \textbf{Medium} & \textbf{Hard} & \textbf{Total} \\ \midrule
\multirow{3}{*}{Qwen-2.5-7b} & Induction & 65.18 & 36.24 & 9.75 & 37.60 \\
 & Automatic & 54.59 & 24.71 & 7.50 & 29.36 \\
 & Abduction+Deduction & 57.88 & 24.71 & 8.00 & 30.64 \\ \midrule
\multirow{3}{*}{Qwen-2.5-72b} & Induction & 79.06 & 49.65 & 17.25 & 49.28 \\
 & Automatic & 74.59 & 44.94 & 13.75 & 45.04 \\
 & Abduction+Deduction & 80.47 & 47.53 & 17.75 & 48.32 \\ \midrule
\multirow{3}{*}{GPT-4o-mini} & Induction & 75.53 & 43.53 & 13.75 & 44.88 \\
 & Automatic & 65.65 & 32.94 & 10.50 & 36.88 \\
 & Abduction+Deduction & 73.41 & 41.65 & 14.00 & 43.60 \\ \bottomrule
\end{tabular}
\caption{LLM performances on List Function dataset in FTG task format.}
\label{tab:list-ftg}
\end{table*}

\begin{table*}[]
\centering
\scriptsize
\begin{tabular}{clcccc}
\toprule
\textbf{Model} & \multicolumn{1}{c}{\textbf{Pipeline}} & \textbf{Easy} & \textbf{Medium} & \textbf{Hard} & \textbf{Total} \\ \midrule
\multirow{3}{*}{Qwen-2.5-7b} & Induction & 21.50 & 16.75 & 10.00 & 16.08 \\
 & Automatic & 36.00 & 31.75 & 19.75 & 29.17 \\
 & Abduction+Deduction & 35.25 & 31.50 & 22.75 & 29.83 \\ \midrule
\multirow{3}{*}{Qwen-2.5-72b} & Induction & 58.50 & 54.25 & 52.00 & 54.92 \\
 & Automatic & 61.25 & 60.00 & 60.00 & 60.42 \\
 & Abduction+Deduction & 60.75 & 62.00 & 63.25 & 62.00 \\ \midrule
\multirow{3}{*}{GPT-4o-mini} & Induction & 63.50 & 56.75 & 39.75 & 53.33 \\
 & Automatic & 53.00 & 47.25 & 43.50 & 47.92 \\
 & Abduction+Deduction & 64.75 & 61.25 & 53.25 & 59.75 \\ \bottomrule
\end{tabular}
\caption{LLM performances on SALT dataset in MCQ task format.}
\label{tab:salt-mcq}
\end{table*}

\begin{table*}[]
\centering
\scriptsize
\begin{tabular}{clcccc}
\toprule
\textbf{Model} & \multicolumn{1}{c}{\textbf{Pipeline}} & \textbf{Easy} & \textbf{Medium} & \textbf{Hard} & \textbf{Total} \\ \midrule
\multirow{3}{*}{Qwen-2.5-7b} & Induction & 37.50 & 15.25 & 2.00 & 18.25 \\
 & Automatic & 29.00 & 17.25 & 6.75 & 17.67 \\
 & Abduction+Deduction & 29.25 & 16.00 & 6.50 & 17.25 \\ \midrule
\multirow{3}{*}{Qwen-2.5-72b} & Induction & 51.25 & 33.50 & 19.50 & 34.75 \\
 & Automatic & 38.00 & 35.25 & 26.75 & 33.33 \\
 & Abduction+Deduction & 33.50 & 30.25 & 30.00 & 31.25 \\ \midrule
\multirow{3}{*}{GPT-4o-mini} & Induction & 66.25 & 25.00 & 17.25 & 36.17 \\
 & Automatic & 34.00 & 25.50 & 20.00 & 26.50 \\
 & Abduction+Deduction & 38.75 & 34.00 & 25.75 & 32.83 \\ \bottomrule
\end{tabular}
\caption{LLM performances on SALT dataset in FTG task format.}
\label{tab:salt-ftg}
\end{table*}

\end{document}